\documentclass[10pt,twocolumn,letterpaper]{article}

\usepackage[pagenumbers]{iccv} %

\def\dsname{MMPR}
\def\modelname{InternVL2-8B-MPO}

\usepackage{booktabs}
\usepackage{multirow}

\usepackage{colortbl}
\usepackage{soul}

\usepackage{marvosym}

\definecolor{mygray}{gray}{.92}
\definecolor{mygreen}{rgb}{0, 0.69, 0.31}

\definecolor{iccvblue}{rgb}{0.21,0.49,0.74}
\usepackage[pagebackref,breaklinks,colorlinks,allcolors=iccvblue]{hyperref}

\def\paperID{8530} %
\def\confName{ICCV}
\def\confYear{2025}

\title{Enhancing the Reasoning Ability of Multimodal Large Language Models \\ via Mixed Preference Optimization}

\author{
\textbf{
    Weiyun Wang$^{2,1}$,
    Zhe Chen$^{3,1}$,
    Wenhai Wang$^{4,1}$,
    Yue Cao$^{3,1}$,
    Yangzhou Liu$^{3,1}$,
}
\\
\textbf{
    Zhangwei Gao$^{1}$,
    Jinguo Zhu$^{1}$,
    Xizhou Zhu$^{5,1}$,
    Lewei Lu$^{6}$,
    Yu Qiao$^{1}$,
    Jifeng Dai$^{5,1}$\textsuperscript{\Letter}
}
\\
$^1$OpenGVLab, Shanghai AI Laboratory,
$^2$Fudan University,
$^3$Nanjing University,
\\
$^4$The Chinese University of Hong Kong,
$^5$Tsinghua University,
$^6$SenseTime Research
\\
\href{https://internvl.github.io/blog/2024-12-20-InternVL-2.5-MPO/}{Project Page}
}

\newcommand\blfootnote[1]{%
\begingroup
\renewcommand\thefootnote{}\footnote{#1}%
\addtocounter{footnote}{-1}%
\endgroup
}

\begin{document}
\maketitle

\blfootnote{{\Letter} Corresponding Author: daijifeng@tsinghua.edu.cn}

\begin{abstract}
Existing open-source multimodal large language models (MLLMs) generally follow a training process involving pre-training and supervised fine-tuning. However, these models suffer from distribution shifts, which limit their multimodal reasoning, particularly in the Chain-of-Thought (CoT) performance.
To address this, we introduce a preference optimization (PO) process to enhance the multimodal reasoning capabilities of MLLMs.
Specifically,
(1) on the data side, we design an automated preference data construction pipeline to create {\dsname}, a high-quality, large-scale multimodal reasoning preference dataset;
and (2) on the model side, we explore integrating PO with MLLMs, developing a simple yet effective method, termed Mixed Preference Optimization (MPO), which boosts multimodal CoT performance. 
Our approach enhances the multimodal reasoning abilities of both InternVL2-8B and InternVL2-76B.
Notably, our model, {\modelname}, achieves an accuracy of 67.0 on MathVista, outperforming InternVL2-8B by 8.7 points and achieving performance comparable to the 10$\times$ larger InternVL2-76B.
We hope this study could inspire further advancements in MLLMs.
Code, data, and model are released in this \href{https://internvl.github.io/blog/2024-12-20-InternVL-2.5-MPO/}{page}.

\end{abstract}

\section{Introduction}
\label{sec:intro}

\begin{figure}[t]
\centering
{\includegraphics[width=\linewidth]{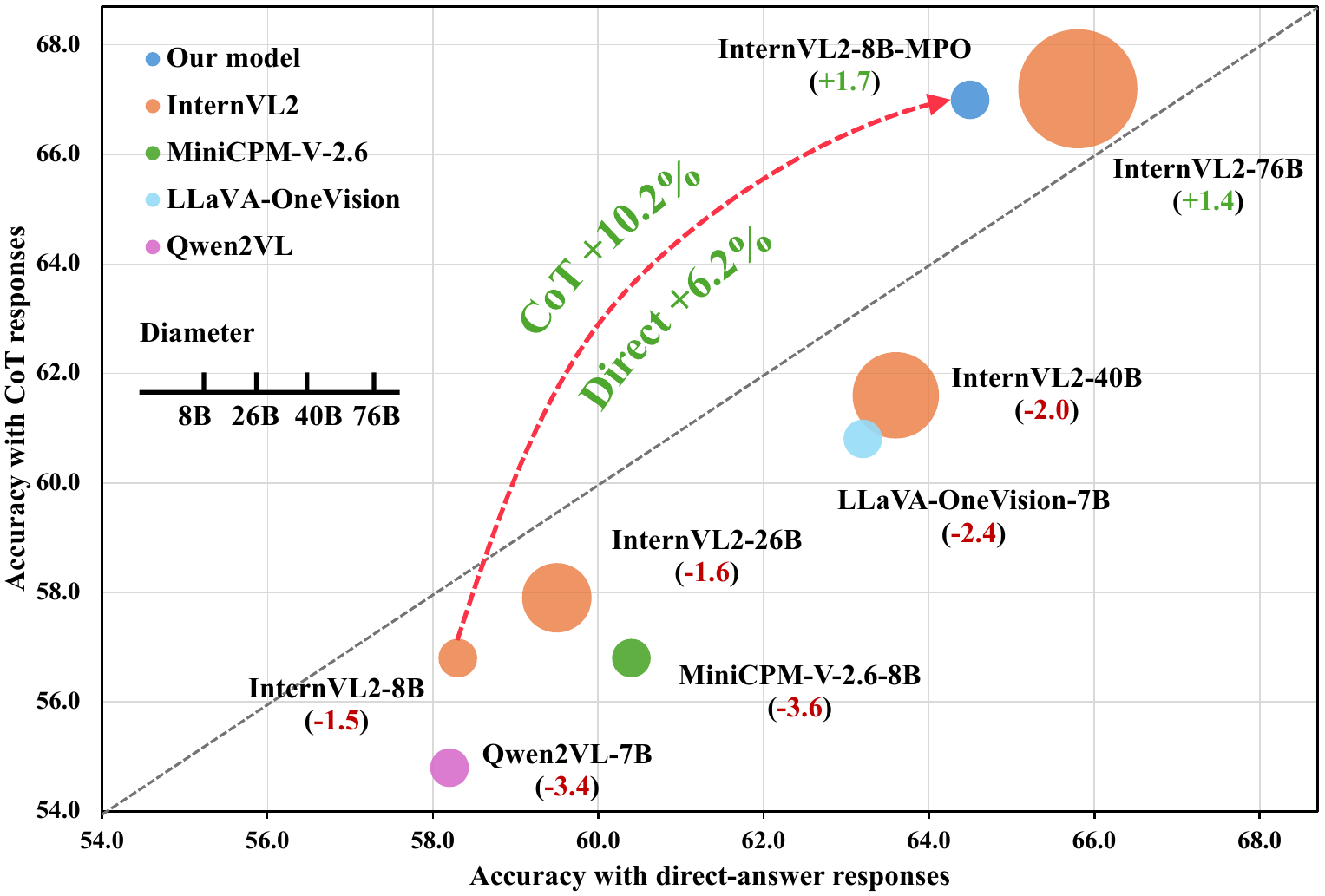}}
\caption{
    \textbf{Open-source model performance on MathVista.}
    The X- and Y-axes represent the accuracy evaluated with direct-answer responses and CoT responses, respectively. The bubble size is positively correlated with the number of model parameters.
    The values in parentheses indicate the performance gap between CoT and direct-answer responses.
    Notably, most open-source models perform worse when answering with CoT.
}
\label{fig:teaser}
\end{figure}

With the remarkable success of large language models (LLMs)~\cite{touvron2023llama,touvron2023llama2,dubey2024llama3,bai2023qwen,2023internlm,cai2024internlm2,brown2020gpt3,openai2023gpt4} in the field of natural language processing, the training paradigm comprising pre-training and supervised fine-tuning (SFT) have also swept the multimodal field, becoming the primary choice for the research and development of multimodal large language models (MLLMs).
Benefiting from the large-scale pre-training corpora~\cite{li2024omnicorpus,wang2023allseeing,schuhmann2022laion5b,thomee2016yfcc100m,zhu2024mmc4,laurenccon2024obelics} and high-quality SFT data~\cite{wang2024allseeingv2,liu2024mminstruct,liu2023llava,chen2024internvl_1_5,instructblip}, a series of open-source MLLMs~\cite{wang2024allseeingv2,chen2024internvl_1_5,liu2023improved,li2023blip2,yao2024minicpm_v,li2024llava_onevision,bai2023qwenvl,wang2024qwen2vl} exhibit strong performance across various domain and tasks, some even achieving results comparable to commercial models such as GPT-4o~\cite{gpt4v} and Gemini~\cite{team2023gemini,reid2024gemini1_5}.

However, open-source MLLMs still exhibit limited reasoning capabilities. 
As shown in Figure~\ref{fig:teaser}, InternVL2-8B \cite{chen2024internvl_1_5} achieves a score of 58.3 on MathVista~\cite{lu2023mathvista}, a benchmark for multimodal reasoning, when using direct answers but drops to 56.8 with Chain-of-Thought (CoT) reasoning, indicating that CoT reasoning actually reduces its performance. 
This decline is commonly observed across open-source MLLMs~\cite{li2024llava_onevision,yao2024minicpm_v,chen2024internvl_1_5,wang2024qwen2vl}.
We attribute this phenomenon primarily to a \textit{distribution shift} introduced by the SFT loss.
Specifically, SFT relies on teacher forcing, where the model is trained to predict the next token based on previous ground-truth tokens. 
However, during inference, models must predict each token based on their own prior outputs, leading to a distribution shift between training and inference.
Since the direct-answer approach requires only brief responses, while CoT reasoning involves generating a long rationale, the distribution shift problem becomes more severe during CoT. This results in models performing worse with CoT reasoning compared to direct-answer responses.

To address the limitations of CoT reasoning in MLLMs, we draw inspiration from recent NLP approaches~\cite{pang2024rpo,lai2024stepdpo,xi2024r3} that use Preference Optimization (PO) techniques to align model outputs with desired reasoning patterns. 
Specifically, methods like Direct Preference Optimization (DPO)~\cite{rafailov2024dpo} allow models to learn from preference signals to generate responses that better align with user requirements, offering the foundation for Reinforcement Learning from Human Feedback (RLHF). While RLHF has been explored for MLLMs primarily to reduce hallucinations~\cite{sun2023llava_rlhf,yu2024rlhf_v,chen2024dress}, its application for enhancing multimodal reasoning remains under-explored.
Building on these insights, we conduct a systematic study on using PO to strengthen the multimodal reasoning capabilities of MLLMs.

Enhancing the multimodal reasoning abilities of MLLMs through PO presents several challenges:
(1) \textit{Limited multimodal reasoning preference data and high annotation cost.} Existing multimodal preference datasets \cite{yu2024rlaif_v,yu2024rlhf_v,sun2023llava_rlhf,li2023silkie,zhang2024spa_vl} primarily address hallucination issues and focus on natural images and perception data, lacking scientific images and reasoning data.
Annotating these types of data requires human annotators to carefully compare the given reasoning processes, making it both time-consuming and costly.
(2) \textit{Lack of open-source methods for improving multimodal reasoning via PO.}
Although previous works have explored fine-tuning MLLMs using feedback from various sources, these models typically exhibit performance gains on hallucination benchmarks,
with little enhancement in general reasoning abilities.
Thus, leveraging PO to improve multimodal reasoning capabilities remains largely under-explored.

This work addresses these challenges from both the data and model sides.
(1) \textit{On the data side,} we design an automated preference data construction pipeline to create {\dsname}, a high-quality, large-scale multimodal reasoning preference dataset.
(2) \textit{On the model side,} we explore various PO methods with MLLMs, introducing a simple yet effective method, termed Mixed Preference Optimization (MPO), which boosts multimodal CoT performance without the requirement for a reward model.

Specifically, we propose a continuation-based pipeline called Dropout Next Token Prediction (DropoutNTP) for samples lacking clear ground truth and a correctness-based pipeline for samples with clear ground truth.
In DropoutNTP, the responses generated by InternVL2 series are considered as positive samples.
For a given chosen response, we truncate it by half and then prompt InternVL2 series to complete the remaining portion of the truncated answer without access to the image input.
This generated completion serves as the rejected answer for the paired sample.
Experimental results in Section~\ref{sec:exp-ablation} demonstrate that this straightforward method achieves comparable performance in reducing hallucinations compared to the divide-and-conquer method proposed in RLAIF-V~\cite{yu2024rlaif_v}.
In the correctness-based pipeline, multiple solutions to each question are sampled from InternVL2 series. Solutions matching the ground truth answer are used as chosen responses, while those that do not are used as rejected responses.

Additionally, we propose the MPO method.
The key insight behind this algorithm is that \textit{an effective PO process should enable the model to learn the relative preference between pairs of responses, the absolute quality of individual responses, and the process for generating preferred responses.}
Compared to previous multimodal PO methods~\cite{yu2024rlaif_v,yu2024rlhf_v,sun2023llava_rlhf,li2023silkie,pi2024bpo,zhang2024spa_vl}, our approach excels in the following aspects:
(1) \textit{Efficient automated data construction pipeline}: Our pipeline enables high-quality preference pair generation at a controlled cost.
(2) \textit{Effectiveness across diverse domains}: Models fine-tuned with our data and approach show superior performance across reasoning, question-answering, and hallucination benchmarks.
(3) \textit{Improvements over SoTA settings}: Our results demonstrate that our method greatly enhances the reasoning abilities of both InternVL2-8B and InternVL2-76B, further highlighting the potential and scalability of our method.

In summary, our main contributions are as follows:

(1) We propose an efficient preference data construction pipeline. Based on this pipeline, we create {\dsname}, a high-quality, large-scale multimodal reasoning preference dataset containing approximately 3 million samples.

(2) We introduce MPO, an effective PO algorithm designed to improve the reasoning abilities of MLLMs. The resulting models, {\modelname} and InternVL2-76B-MPO, exhibit enhanced multimodal reasoning ability compared to their baseline models before MPO.

(3) We conduct extensive experiments to explore practical approaches for improving multimodal reasoning via PO. Results show that PO significantly improves reasoning abilities over SFT.
Notably, the proposed {\modelname} achieves an accuracy of 67.0 on MathVista~\cite{lu2023mathvista}, outperforming InternVL2-8B by 8.7 points and achieving performance comparable to the 10$\times$ larger InternVL2-76B.

\section{Related Work}
\label{sec:related_work}

\noindent\textbf{Multimodal Large Language Models.}
With advancements in LLMs, significant progress has also been made in MLLMs.
To leverage the abilities of pre-trained LLMs~\cite{cai2024internlm2,bai2023qwen,dubey2024llama3} and Vision Foundation Models (VFMs)~\cite{rebuffi2017learning,chen2023internvl}, a series of works~\cite{2023interngpt,li2022blip,li2023blip2,liu2023llava,wang2024qwen2vl,wang2023allseeing,chen2024internvl_1_5,wang2024mmniah} employ a connector to align their latent space, achieving promising performance at a controllable cost.
Besides, another series of works~\cite{alayrac2022flamingo,wang2023cogvlm,tian2024mminterleaved,dubey2024llama3} extend pre-trained LLMs with additional fusion layers for vision features, reducing the number of visual tokens required by LLMs while introducing extra training costs.
Recently, there have been explorations into vision encoder-free architectures~\cite{fuyu-8b,lin2024moma,team2024chameleon,luo2024mono_internvl,wang2024emu3}, which consist of a single transformer model that jointly processes visual and textual information without a separate encoder.
In addition to exploring model architectures, recent works~\cite{liu2024mminstruct,li2024omnicorpus,zhang2024mavis,yang2024mathglm_v,gao2023gllava,wang2024allseeingv2} also try to construct high-quality training data to improve multimodal reasoning abilities.
Despite these advancements, MLLMs typically rely on a training paradigm comprising pre-training and supervised fine-tuning, which suffers from the curve of distribution shift and exhibits limited multimodal reasoning abilities.
In this work, we conduct a systematic study on using preference optimization to enhance the multimodal reasoning ability of MLLMs.

\noindent\textbf{Preference Optimization.}
Preference optimization (PO) is a crucial technique for advancing LLMs and MLLMs. Specifically, Reinforcement Learning from Human Feedback (RLHF) uses human preferences as a reward signal to fine-tune models, aligning them with human preferences. InstructGPT~\cite{ouyang2022instructgpt} employs a reward model as a proxy for human preferences and maximizes this reward via the PPO algorithm~\cite{schulman2017ppo}, improving the model's ability to follow user intent and become more helpful, honest, and harmless (3H). 
PPO-Max~\cite{zheng2023ppomax_part1,wang2024ppomax_part2} carefully explores the implementation details of PPO, proposing a more stable version of the algorithm.
Additionally, DPO~\cite{rafailov2024dpo} proposes an efficient PO algorithm based on the Bradley-Terry model~\cite{bradley1952bt_rm}, removing the need for an explicit reward model.
Subsequent works~\cite{azar2024ipo,dong2024iterdpo,lai2024stepdpo,liu2023rso,chowdhury2024robustdpo,hong2024orpo,gorbatovski2024tr_dpo} have further analyzed and refined this method from various perspectives.
In natural language processing, a series of works~\cite{pang2024rpo,lai2024stepdpo} have explored how to leverage PO to enhance reasoning ability.
In the multimodal field, however, most methods~\cite{yu2024rlaif_v,yu2024rlhf_v,zhang2024spa_vl,sun2023llava_rlhf,li2023silkie} primarily focus on reducing hallucination, leaving the potential for PO to improve multimodal reasoning ability under-explored. 
This work demonstrates that PO not only mitigates hallucinations but also strengthens multimodal reasoning abilities, highlighting its broader applicability in MLLM development.

\section{Scalable Multimodal Preference Dataset Generation}
\label{sec:methods-ds}

To address the scarcity of multimodal preference data, we introduce a scalable data construction pipeline.
Based on this pipeline, we construct a million-level \textbf{M}ulti\textbf{M}odal \textbf{PR}eference dataset ({\dsname}).

\subsection{Data Engine}
\label{sec:methods-ds-construction}

\textbf{Definition.} 
Each data sample in our {\dsname} consists of an image $I \in \mathcal{I}$, an instruction $x \in \mathcal{X}$, a chosen response $y_c \in \mathcal{Y}_p$, and a rejected response $y_r \in \mathcal{Y}_n$, where $y_c$ is preferable to $y_r$.
The image sets $\mathcal{I}$ and instruction sets $\mathcal{X}$ are collected from existing datasets. $\mathcal{Y}_p$ and $\mathcal{Y}_n$ represent the positive and negative response set, respectively.
Given a certain image $I$ and instruction $x$, we sample the candidate response $y$ from an initial instruction model $M_0$ as follows:
\begin{equation}
    y \sim M_0(y \mid x, I),
    \label{equ:method-response}
\end{equation}
where $M_0(y \mid x, I)$ represents the response distribution of $M_0$ conditioned on image $I$ and instruction $x$.

\noindent\textbf{For instructions with clear ground truths}, the model is prompted to first provide the reasoning process and then give the final answer in the format like ``\texttt{Final Answer: ***}''.
Responses matching the ground truth answer constitute the positive set $\mathcal{Y}_p$, while those that do not match make up the negative set $\mathcal{Y}_n$. Additionally, responses that fail to provide a clear final answer are also merged into $\mathcal{Y}_n$.
Given these responses labeled as positive or negative, we build the preference pairs by selecting a chosen response $y_c$ from $\mathcal{Y}_p$ and a negative response $y_r$ from $\mathcal{Y}_n$.

\noindent\textbf{For instructions without clear ground truths}, we propose a simple yet effective method: Dropout Next-Token Prediction (Dropout NTP).
Specifically, we directly consider all responses generated from Equation~\ref{equ:method-response} as positive set $\mathcal{Y}_p$.
To generate the negative set $\mathcal{Y}_n$, we sample a response $y$ from $\mathcal{Y}_p$ and drop the last half of this response. The model is required to complete the remained response as follows:
\begin{equation}
    \tilde{y}_{\geq j} \sim M_0(\tilde{y}_{\geq j} \mid x, y_{<j}),
    \label{equ:method-completation}
\end{equation}
where $y_{<j}$ and $y_{\geq j}$ is the remained part and truncated part of $y$, respectively. $\tilde{y}_{\geq j}$ is the completion of $y_{<j}$ without the image input.
The original response $y=\left[y_{<j}, y_{\geq j}\right]$ serves as the chosen response $y_c$ and the completed response $\tilde{y}=\left[y_{<j}, \tilde{y}_{\geq j}\right]$ serves as the rejected response $y_r$.
It is worth noting that while the responses generated by $M_0$ may not be perfect, the completions generated without the image input will introduce more hallucinations than those generated with the image input.
Therefore, the partial order relationship between $y$ and $\tilde{y}$ holds true.

\noindent\textbf{Compared with previous methods}, 
our data engine is as effective as the more complex divide-and-conquer method proposed in RLAIF-V~\cite{yu2024rlaif_v} (see the experimental results in Section~\ref{sec:exp-ablation-rlaifv}), while more efficient.
Taking data generation for M3CoT as an example, our pipeline incurs a token cost of {571.2} per preference pair, compared to {992.7} tokens for the divide-and-conquer approach used in RLAIF-V. Thus, the cost of our pipeline is only {57.5}\% of that of RLAIF-V.
Additionally, a comparison with other recent data  pipelines~\cite{zhou2024povid,deng2024stic,mukherjee2023seva} is also presented in Section~\ref{sec:exp-ablation-rlaifv}.

\subsection{Multimodal Preference Dataset}
\label{sec:methods-ds-ds}

\begin{figure*}[t]
\centering
\begin{subfigure}[b]{\textwidth}
    {\includegraphics[width=\textwidth]{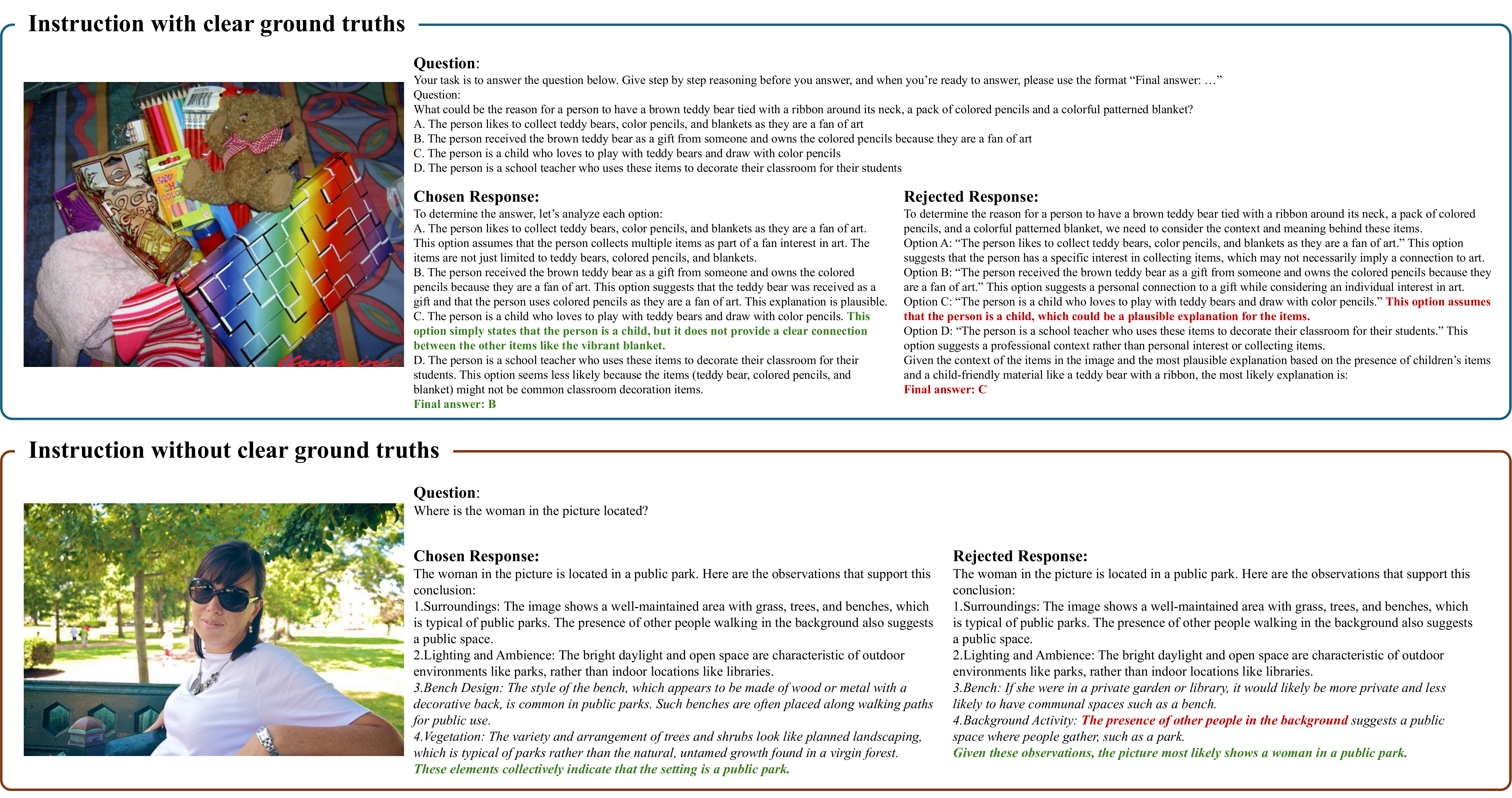}}
\end{subfigure}
\caption{
    \textbf{Data examples in {\dsname}.}
    For instructions with clear ground truths, we propose a correctness-based pipeline, which samples multiple solutions and considers those with correct answers as chosen responses and those with incorrect answers as rejected responses.
    For instructions without clear ground truths, we propose DropoutNTP to generate rejected responses. Differences between the chosen and rejected responses are emphasized in \textit{italicized text}.
    \textcolor{red}{Red} highlights incorrect responses.
}
\label{fig:data-examples}

\vspace{-3mm}

\end{figure*}

\noindent\textbf{Dataset Statistics.}
Using this pipeline, we build a large-scale multimodal preference dataset, {\dsname}.
Data examples are presented in Figure~\ref{fig:data-examples}. See more examples in the Appendix.
This dataset comprises approximately 750K samples without clear ground truths and 2.5M samples with clear ground truths.
For samples without clear ground truths, each instruction averages 25.0 tokens, while the chosen and rejected responses average 211.4 and 171.2 tokens, respectively. The longest chosen and rejected responses consist of 1,342 and 1,642 tokens, respectively, whereas the shortest chosen and rejected responses contain 20 and 17 tokens, respectively.
For samples with clear ground truths, the average instruction length is 79.5 tokens, with the chosen and rejected responses averaging 300.0 and 350.5 tokens, respectively. The longest chosen and rejected responses are composed of 2,018 and 4,097 tokens, while the shortest responses contain 32 and 33 tokens, respectively.

\noindent\textbf{Data Source.} As shown in Table~\ref{tab:data-source}, to ensure the diversity of instructions and images, we collect samples from diverse domains, including general visual question answering (VQA)~\cite{goyal2017vqav2,hudson2019gqa,marino2019okvqa,lu2021iconqa}, science~\cite{kembhavi2016ai2d,chen2024m3cot,lu2022scienceqa}, chart~\cite{masry2022chartqa,kafle2018dvqa,chang2022mapqa},  mathematics~\cite{lindstrom2022clevrmath,seo2015geos,cao2022geoqa_plus,lu2021geometry3k,kazemi2023geomverse,gao2023gllava}, OCR~\cite{mathew2022infographicvqa,singh2019textvqa,biten2019stvqa,huang2019sroie,mishra2019ocrvqa}, and document~\cite{clark2017docqa}.
Notably, when constructing open-ended samples, we collect instructions from all the data sources mentioned above and prompt the model to answer the original question without additional requirements.
On the other side, when building samples through the correctness-based pipeline, we exclude questions from general VQA and document sources, as verifying the correctness of the generated answers using heuristic rules is challenging for datasets in these domains.
For example, the ground truths in VQAv2~\cite{goyal2017vqav2} consist of a single word or phrase, which may lead to false-negative responses when the model outputs a complete sentence or a synonym as the final answer.
Such false-negative responses can negatively impact training effectiveness.

\begin{table}[t]
\centering
\footnotesize
\setlength\tabcolsep{3pt}
\renewcommand{\arraystretch}{0.85}

\begin{tabular}{l|l}
\toprule
Task        & Dataset                                         \\ \midrule
General VQA & VQAv2~\cite{goyal2017vqav2}, GQA~\cite{hudson2019gqa}, OKVQA~\cite{marino2019okvqa}, IconQA~\cite{lu2021iconqa}                       \\
\rowcolor{mygray}
Science     & AI2D~\cite{kembhavi2016ai2d}, ScienceQA~\cite{lu2022scienceqa}, M3CoT~\cite{chen2024m3cot}                          \\
Chart       & ChartQA~\cite{masry2022chartqa}, DVQA~\cite{kafle2018dvqa}, MapQA~\cite{chang2022mapqa}                            \\
\rowcolor{mygray}
Mathematics & \begin{tabular}[c]{@{}l@{}}GeoQA+~\cite{cao2022geoqa_plus}, CLEVR-Math~\cite{lindstrom2022clevrmath}, Geometry3K~\cite{lu2021geometry3k}, \\GEOS~\cite{seo2015geos}, GeomVerse~\cite{kazemi2023geomverse}, Geo170K~\cite{gao2023gllava}\end{tabular} \\
OCR         & \begin{tabular}[c]{@{}l@{}}OCRVQA~\cite{mishra2019ocrvqa}, InfoVQA~\cite{mathew2022infographicvqa}, TextVQA~\cite{singh2019textvqa}, \\STVQA~\cite{biten2019stvqa}, SROIE~\cite{huang2019sroie}\end{tabular}                  \\
\rowcolor{mygray}
Document    & DocVQA~\cite{mathew2021docvqa}                                          \\ \bottomrule
\end{tabular}

\caption{
\textbf{Datasets used to build our preference dataset.}
}
\label{tab:data-source}

\vspace{-3mm}

\end{table}

\section{Improved Multimodal Large Language Model with Preference Optimization}
\label{sec:methods-training}

To enhance the multimodal reasoning capabilities of MLLMs, we propose mixed preference optimization (MPO), a method that blends supervised fine-tuning (SFT) loss with various preference optimization losses to enhance training effectiveness. Additionally, we investigate different Chain-of-Thought (CoT) approaches with multimodal input to improve reasoning performance.

\subsection{Mixed Preference Optimization}
\label{sec:methods-loss}

We observed that when MLLMs are trained on large-scale preference datasets using direct preference optimization (DPO), they might fail to generate reasonable rationales and produce repetitive responses, as shown in Section~\ref{sec:suppl-qualitative}. 
This phenomenon aligns with the analysis presented in Smaug~\cite{pal2024smaug}.
To address this issue, we introduce the MPO in this work, aiming to learn the relative preference between pairs of responses, the absolute quality of individual responses, and the process for generating preferred responses.

\noindent\textbf{Training Objective.} MPO is defined as a combination of preference loss $\mathcal{L}_{p}$, quality loss $\mathcal{L}_{q}$, and generation loss $\mathcal{L}_{g}$, which can be formulated as follows:
\begin{equation}
    \mathcal{L}=
    w_{p} \mathcal{L}_{p}
    +
    w_{q} \mathcal{L}_{q}
    +
    w_{g} \mathcal{L}_{g}
    ,
    \label{eqn:final_loss}
\end{equation}
where $w_{*}$ represents the weight assigned to each loss component.
In this work, we empirically compare different variants of preference loss~\cite{mitchell2023cdpo,chowdhury2024robustdpo,chen2024nca,jung2024bco,wu2024sppo,melnyk2024aot,azar2024ipo,liu2023rso,rafailov2024dpo,hong2024orpo}.
Based on the experimental results, we use DPO~\cite{rafailov2024dpo} as our preference loss and BCO~\cite{jung2024bco} as our quality loss.

\noindent\textbf{Preference Loss.} The DPO~\cite{rafailov2024dpo} serves as the preference loss to enable the model to learn the relative preference between chosen and rejected responses.
DPO eliminates the requirement of training an explicit reward model based on the assumption of the Bradley-Terry model~\cite{bradley1952bt_rm} and optimizes the following loss function:
{\small
\begin{equation}
    \mathcal{L}_p=
    -\log \sigma\left(\beta \log \frac{\pi_\theta\left(y_c \mid x\right)}{\pi_0\left(y_c \mid x\right)}-\beta \log \frac{\pi_\theta\left(y_r \mid x\right)}{\pi_0\left(y_r \mid x\right)}\right),
    \label{eqn:dpo_loss}
\end{equation}
}%
where $\beta$ is the KL penalty coefficient,
and $x$, $y_c$, and $y_r$ are user query, chosen response, and rejected response, respectively.
The policy model $\pi_\theta$ is initialized from model $\pi_0$.

\noindent\textbf{Quality Loss.} The BCO loss~\cite{jung2024bco} is employed as the quality loss, which helps the model to understand the absolute quality of individual responses. This algorithm trains a binary classifier, where the logit serves as a reward and effectively maps the chosen response to 1 and the rejected response to 0.
The loss function is defined as:
\begin{equation}
    \mathcal{L}_{q}=
    \mathcal{L}_{q}^+
    +
    \mathcal{L}_{q}^-,
\end{equation}
where $\mathcal{L}_{q}^+$ and $\mathcal{L}_{q}^-$ represent the loss for chosen and rejected responses, respectively.
They are calculated independently, requiring the model to differentiate the absolute quality of individual responses.
The loss terms are given by:
\begin{equation}
    \mathcal{L}_{q}^+=
    -\log \sigma\left(
        \beta \log \frac{\pi_\theta\left(y_c \mid x\right)}{\pi_0\left(y_c \mid x\right)}
        - \delta
    \right),
\end{equation}
\begin{equation}
    \mathcal{L}_{q}^-=
    - 
    \log \sigma\left(
    -\left(
        \beta \log \frac{\pi_\theta\left(y_r \mid x\right)}{\pi_0\left(y_r \mid x\right)}
        - \delta
    \right) \right),
\end{equation}
where $\delta$ represents the reward shift, calculated as the moving average of previous rewards to stabilize training.

\noindent\textbf{Generation Loss.} The SFT loss is used as the generation loss to help the model learn the generation process of preferred responses.
The loss function is defined as:
\begin{equation}
    \mathcal{L}_{g}=-\frac{\log\pi_\theta\left(y_c \mid x\right)}{\left| y_c \right|}.
    \label{eqn:sft_loss}
\end{equation}

\subsection{Chain-of-Thought with Multimodal Input}
\label{sec:methods-cot}

During the data sampling process, we require the model to provide a detailed CoT reasoning process instead of directly answering the final answer.
For most samples, we sample the responses using the prompt shown in the bottom case of Figure~\ref{fig:data-examples}, which requires the model to perform a step-by-step analysis.
Considering that MLLMs involve non-textual inputs, we further introduce the following CoT methods:
(1) \textbf{Background Knowledge-based CoT:} The model first introduces relevant background knowledge related to the problem or image, followed by reasoning steps and the final answer. This approach is applied to samples from the science domain.
(2) \textbf{Visual Content-based CoT:} The model begins by analyzing the visual contents in the image, then proceeds with reasoning and the final answer. This method is used for samples from chart, OCR, and document domains.
(3) \textbf{Grounded CoT:} The model generates a text response while simultaneously linking all referenced objects in the response to corresponding regions in the image. This approach is applied to general VQA domain samples.

Responses generated by these above CoT methods are mixed with those sampled using the prompt shown in the bottom case of Figure~\ref{fig:data-examples}.
These approaches not only effectively integrate multimodal information into the reasoning process but also enhance data diversity.
Furthermore, including the background knowledge and visual contents at the start of responses also improves the quality of the negative responses generated by DropoutNTP, preventing a significant quality gap between positive and negative samples that could reduce training effectiveness.

\section{Experiments}
\label{sec:exp}

\subsection{Main Results}
\label{sec:exp-main}

\begin{table*}[t]
\centering
\small
\setlength\tabcolsep{2.5pt}
\renewcommand{\arraystretch}{0.87}

\begin{tabular}{lccccccccc}
\toprule
\textbf{Model Name}      & \textbf{M3CoT} & \textbf{MMMU} & \textbf{MathVista} & \textbf{MathVision} & \textbf{MathVerse-VO} & \textbf{WeMath} & \textbf{LogicVista} & \textbf{MMVet} & \textbf{MMHal} \\ \midrule
\multicolumn{10}{c}{\textit{Proprietary Models}}                                                                                                                                                            \\ \midrule
Gemini-1.5-Pro~\cite{reid2024gemini1_5}           & -              & -             & 63.9               & 19.2                & -                     & -               & -                   & -              & -                   \\
GPT-4o~\cite{gpt4o}                   & 64.3           & 70.7          & 63.8               & 30.4                & 40.6                  & 45.8            & 52.8                & 69.1           & 4.0                 \\
GPT-4o-Mini~\cite{gpt4o}              & 61.9           & -             & 52.4               & 27.3                & -                     & -               & -                   & 66.9           & 3.6                 \\ \midrule
\multicolumn{10}{c}{\textit{Open Source Models}}                                                                                                                                                            \\ \midrule
LLaVA-1.5-13B~\cite{liu2023improved}            & 39.5           & 35.7          & 27.6               & 11.1                & 11.4                  & 1.4             & 7.0                 & 36.3           & 2.4                 \\
Qwen2-VL-7B~\cite{wang2024qwen2vl}              & 57.8           & 53.7          & 58.2               & 21.1                & 19.2                  & 11.0            & 22.3                & 60.6           & 3.4                 \\
MiniCPM-V-2-6-8B~\cite{yao2024minicpm_v}         & 56.0           & 49.8          & 60.6               & 23.4                & 18.9                  & 16.4            & 27.5                & 57.4           & 3.6                 \\
LLaVA-OneVision-7B~\cite{li2024llava_onevision}       & 52.3           & 47.9          & 63.2               & 18.4                & 18.3                  & 9.0             & 20.9                & 51.4           & 3.1                 \\
\midrule
\multicolumn{10}{c}{\textit{InternVL Models}}                                                                                                                                   \\
\midrule
InternVL2-8B~\cite{chen2024internvl_1_5}             & 59.3           & 51.2          & 58.3               & 20.4                & 20.4                  & 20.2            & 33.6                & 54.2           & 3.3                 \\
\rowcolor{mygray}
InternVL2-8B-MPO (ours)  & 79.2           & 54.0          & 67.0               & 25.7                & 25.0                  & 28.4            & 39.8                & 56.2           & 3.5                 \\
InternVL2-26B~\cite{chen2024internvl_1_5}            & 58.2           & 50.7          & 59.4               & 23.4                & 19.5                  & 18.8            & 34.0                & 62.1           & 3.7                 \\
InternVL2-40B~\cite{chen2024internvl_1_5}            & 63.6           & 55.2          & 63.7               & 21.4                & 23.5                  & 22.2            & 39.8                & 65.5           & 3.9                 \\
InternVL2-76B~\cite{chen2024internvl_1_5}            & 65.4           & 58.3          & 67.5               & 23.7                & 23.7                  & 32.1            & 45.6                & 65.7           & 3.8                 \\
\rowcolor{mygray}
InternVL2-76B-MPO (ours) & 82.0           & 64.4          & 70.9               & 30.6                & 37.7                  & 39.4            & 49.0                & 69.5           & 4.2                 \\ \bottomrule
\end{tabular}

\caption{
\textbf{Results on multimodal benchmarks.}
M3CoT~\cite{chen2024m3cot} and MMMU~\cite{yue2023mmmu} are multidisciplinary reasoning benchmarks.
MathVista~\cite{lu2023mathvista}, MathVision~\cite{wang2024mathvision}, MathVerse~\cite{zhang2024mathverse}, and WeMath~\cite{qiao2024wemath} are mathematics benchmarks. For MathVerse, we report the performance on Vision-Only (VO) split.
LogicVista~\cite{xiao2024logicvista} is a logical reasoning benchmark.
Additionally, MMVet~\cite{yu2023mmvet} and MMHal~\cite{sun2023llava_rlhf} are designed for general VQA and hallucination evaluation, respectively.
Our {\modelname} demonstrates superior performance compared to InternVL2-8B across multimodal reasoning, VQA, and hallucination evaluation benchmarks. 
Notably, both InternVL2-8B-MPO and InternVL2-76B-MPO exhibit significant performance improvements over their counterparts before MPO.
}
\label{tab:exp-main}

\end{table*}

In this section, we compare our models with leading MLLMs on multimodal reasoning~\cite{chen2024m3cot,yue2023mmmu,lu2023mathvista,wang2024mathvision,zhang2024mathverse,qiao2024wemath,xiao2024logicvista}, complex Visual Question Answering (VQA)~\cite{yu2023mmvet}, and hallucination evaluation~\cite{sun2023llava_rlhf} tasks.

\noindent\textbf{Benchmarks.}
We evaluate the reasoning abilities of MLLMs across seven benchmarks, including M3CoT~\cite{chen2024m3cot}, MMMU~\cite{yue2023mmmu}, MathVista~\cite{lu2023mathvista}, MathVision~\cite{wang2024mathvision}, MathVerse~\cite{zhang2024mathverse}, WeMath~\cite{qiao2024wemath}, and LogicVista~\cite{xiao2024logicvista}.
The evaluation samples include subject-based, mathematical, and logical reasoning problems.
We report the overall accuracy for these benchmarks. For MathVerse, we report the performance on the Vision-Only split.
Additionally, we evaluate the general VQA abilities and hallucinations of MLLMs using MMVet~\cite{yu2023mmvet} and MMHalBench~\cite{sun2023llava_rlhf}, respectively.

\noindent\textbf{Results.}
As shown in Table~\ref{tab:exp-main}, {\modelname} and InternVL2-76B-MPO achieve superior performance across all benchmarks compared to their counterparts before MPO, particularly excelling in multimodal reasoning tasks.
On the MathVista benchmark, {\modelname} achieves an accuracy of 67.0\%, outperforming InternVL2-8B by 8.7 points and achieving performance comparable to the 10$\times$ larger InternVL2-76B.
Similarly, on the MathVision benchmark, InternVL2-76B-MPO achieves an accuracy of 30.6\%, establishing a new state-of-the-art performance.
These results highlight the effectiveness of our preference optimization approach in enhancing multimodal reasoning capabilities.
Furthermore, our models also show superior performance over their pre-MPO counterparts on VQA~\cite{yu2023mmvet} and hallucination~\cite{sun2023llava_rlhf} benchmarks, indicating that their general abilities are also improved, benefiting from enhanced reasoning abilities and mitigated hallucinations.
Notably, InternVL2-76B-MPO exhibits a more significant performance improvement over InternVL2-76B than the improvement observed in the 8B model, demonstrating the scalability of our method.

\subsection{Ablation Study}
\label{sec:exp-ablation}

In this section, we present ablation studies to analyze the effects of preference optimization and SFT on multimodal reasoning abilities.
Additionally, we compare our proposed DropoutNTP method with recent related data pipelines~\cite{mukherjee2023seva,zhou2024povid,deng2024stic} and the divide-and-conquer approach from RLAIF-V~\cite{yu2024rlaif_v}, demonstrating the effectiveness of our approach.
Furthermore, we conduct extensive experiments to analyze the effects of different preference optimization algorithms.
We also present analysis of the effects on text-only performance.

\subsubsection{Comparison between MPO and SFT}
\label{sec:exp-ablation-rl-and-sft}

\begin{table}[t]
\centering
\footnotesize
\setlength\tabcolsep{3pt}
\renewcommand{\arraystretch}{0.9}

\begin{tabular}{lccccc}
\toprule
\textbf{Model Name}               & \textbf{Setting} & \textbf{M3CoT} & \textbf{MathVista} & \textbf{MMVet} & \textbf{POPE} \\ \midrule
\multirow{2}{*}{InternVL2-8B}     & Direct           & 59.3           & 58.3               & 54.2           & 86.9          \\
                                  & CoT              & 57.0           & 56.8               & 54.7           & 82.9          \\ \midrule
\multirow{2}{*}{InternVL2-8B-SFT} & Direct           & 63.9           & 62.7               & 54.7           & 86.5          \\
                                  & CoT              & 67.8           & 64.2               & 53.8           & 84.0          \\ \midrule
\multirow{2}{*}{{\modelname}} & Direct           & 77.2           & 64.5               & 55.1           & 87.0          \\
                                  & CoT              & 79.2           & 67.0               & 56.2           & 88.1          \\ \bottomrule
\end{tabular}

\caption{\textbf{Results of models trained with SFT and MPO.} The SFT training data consists of the chosen responses from the preference pairs used in MPO training. In the Direct setting, the model is prompted to provide the answer directly, while in the CoT setting, the model is instructed to answer with detailed rationales.}
\label{tab:ablation-sft-rl}

\end{table}

To compare the impact of MPO and SFT on improving multimodal reasoning ability, we use the chosen responses in {\dsname} as SFT data to fine-tune InternVL2-8B.
As shown in Table~\ref{tab:ablation-sft-rl}, the results indicate that the model trained with MPO consistently outperforms that trained with SFT across all benchmarks.
For example, the MPO-trained model achieves a score of 79.2 on the multimodal reasoning benchmark M3CoT, surpassing its SFT counterpart by 11.4 points.
Furthermore, the MPO-trained model also performs better on the general benchmark~\cite{yu2023mmvet} and the hallucination benchmark~\cite{li2023pope}.
Notably, the SFT-trained model performs worse with CoT responses than with direct-answer responses on MMVet and POPE, demonstrating that SFT alone is insufficient to enhance multimodal CoT abilities.
These results demonstrate that while SFT provides moderate improvement, preference optimization is more effective in improving the overall performance of the model.

\subsubsection{Comparison with Different Data Pipelines}
\label{sec:exp-ablation-rlaifv}
Here, we compare our proposed Dropout NTP method with SeVa~\cite{mukherjee2023seva}, POVID~\cite{zhou2024povid}, STIC~\cite{deng2024stic}, and the divide-and-conquer approach from RLAIF-V~\cite{yu2024rlaif_v}. To ensure a fair comparison, we use the same prompts and chosen responses as in RLAIF-V and replace the rejected responses with those generated by different data pipelines.
Following RLAIF-V, we report the hallucination rates in response-level (Resp.) and mention-level (Ment.) for Object HalBench~\cite{rohrbach2018objhal} and overall score and hallucination rates (Hall.) for MMHal-Bench~\cite{sun2023llava_rlhf}.

As shown in Table~\ref{tab:ablation-rlaifv}, the model trained with our data achieves performance comparable to that of the model trained with RLAIF-V, demonstrating the effectiveness of our method.
Specifically, the response-level hallucination rate of the model trained with our data on Object HalBench is {7.6}, compared to {7.3} for its counterpart. Besides, this model achieves a score of {3.6} on the MMHal-Bench, compared to {3.5} for its counterpart.
Note that our method requires the model to generate only a single continuation for each sample, while RLAIF-V requires the model to decompose the response into atomic claims and then verify each one individually. Therefore, our method is more efficient.
A quantitative analysis is provided in Section~\ref{sec:methods-ds-construction}.

Furthermore, DropoutNTP outperforms SeVa, POVID, and STIC. We manually review the generated samples and find that 
\textit{responses generated by DropoutNTP are stronger negative samples, which leads to better training effectiveness.}
We attribute this to the fact that most open-source MLLMs are trained on text-only samples and those paired with clean images.
Therefore, corrupted images are out-of-domain inputs for these models, and responses generated conditioned on corrupted images are much worse than those generated with text-only input.

\begin{table}[t]
\centering
\small
\setlength\tabcolsep{2.5pt}
\renewcommand{\arraystretch}{0.9}

\begin{tabular}{lcccc}
\toprule
\multirow{2}{*}{\textbf{Method}} & \multicolumn{2}{c}{\textbf{Object HalBench}} & \multicolumn{2}{c}{\textbf{MM HalBench}} \\ \cmidrule(l){2-5}
                                 & \textbf{Resp. (↓)}    & \textbf{Ment. (↓)}   & \textbf{Score}    & \textbf{Hall. (↓)}   \\ \midrule
InternVL2-8B                     & 18.4                  & 8.7                  & 3.3               & 40.6                 \\
POVID~\cite{zhou2024povid}                            & 10.5                  & 5.5                  & 3.4               & 38.5                 \\
STIC~\cite{deng2024stic}                             & 9.4                   & 5.3                  & 3.5               & 37.5                 \\
SeVa~\cite{mukherjee2023seva}                             & 9.1                   & 4.6                  & 3.4               & 39.6                 \\
RLAIF-V~\cite{yu2024rlaif_v}                          & 7.3                   & 3.9                  & 3.5               & 32.3                 \\
\rowcolor{mygray}
DropoutNTP (ours)             & 7.6                   & 4.1                  & 3.6               & 31.3                 \\ \bottomrule
\end{tabular}

\caption{
\textbf{Comparison of DropoutNTP and other data construction pipelines.} We replace the negative samples in RLAIF-V with responses generated using different pipelines.
}
\label{tab:ablation-rlaifv}

\end{table}

\subsubsection{Effects of Optimization Algorithms}
\label{sec:exp-ablation-algo}

\begin{figure*}[t]
\centering
{\includegraphics[width=\linewidth]{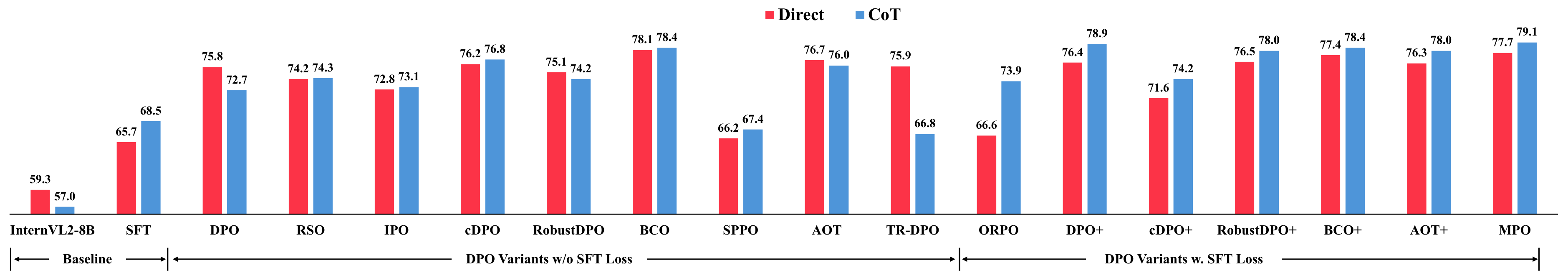}}
\caption{
    \textbf{Results of models trained with different preference optimization algorithms on M3CoT.}
    The algorithm X extended with the SFT loss is called X+ for brevity. For instance, DPO+ denotes the combination of DPO loss and SFT loss.
}
\label{fig:ablation-algo}
\end{figure*}

\begin{table*}[t]
\centering
\footnotesize
\setlength\tabcolsep{2.5pt}

\begin{tabular}{lcccccccccccccc}
\toprule
\textbf{Setting} & \textbf{MMLU} & \textbf{Gaokao} & \textbf{TriviaQA} & \textbf{NQ} & \textbf{C3} & \textbf{Race-h} & \textbf{BBH} & \textbf{GSM8K} & \textbf{Math} & \textbf{TheoremQA} & \textbf{IFEval} & \textbf{HumanEval} & \textbf{MBPP} & \textbf{Average} \\ \midrule
Baseline         & 73.2          & 75.0            & 62.0              & 28.1        & 94.2        & 90.8              & 72.7         & 75.6           & 39.5          & 15.6               & 52.3            & 69.5               & 58.8                     & 62.1             \\
SFT              & 71.8          & 74.4            & 63.7              & 28.2        & 94.3        & 90.6              & 72.1         & 75.5           & 40.1          & 15.8               & 53.6            & 68.3               & 58.0                     & 62.0             \\
MPO              & 71.0          & 74.8            & 64.2              & 29.3        & 94.2        & 90.6              & 71.8         & 75.0           & 40.4          & 20.8               & 56.4            & 68.9               & 61.5                     & 63.0             \\ \bottomrule
\end{tabular}

\caption{
\textbf{Results on text-only benchmarks.}
The model fine-tuned through MPO exhibits superior overall performance on text-only tasks compared to the baseline model and its SFT counterpart, particularly on TheoremQA and IFEval.
}
\label{tab:ablation-text-only-performance}

\end{table*}

Here, we empirically compare the effectiveness of different optimization algorithms, including
(1) \textbf{DPO}~\cite{rafailov2024dpo}, which directly fine-tunes the model on an offline preference dataset without explicitly constructing a reward function.
(2) \textbf{RSO}~\cite{liu2023rso}, which applies a hinge loss on the normalized likelihood instead of the sigmoid loss used in DPO.
(3) \textbf{IPO}~\cite{azar2024ipo}, which introduces a modified loss function to address overfitting in DPO by averaging log-likelihoods and controlling the gap between chosen and rejected completions via a beta parameter.
(4) \textbf{cDPO}~\cite{mitchell2023cdpo}, which is a modification of the DPO loss that accounts for potential label noise in preference data.
(5) \textbf{RobustDPO}~\cite{chowdhury2024robustdpo}, which provides an unbiased estimate of the DPO loss designed to handle preference noise in data. Similar to cDPO, it assumes that labels are noisy with a certain probability.
(6) \textbf{BCO}~\cite{jung2024bco}, which introduces a binary classifier trained to output logits used as reward values.
(7) \textbf{SPPO}~\cite{wu2024sppo}, which iteratively pushes chosen rewards toward 1/2 and rejected rewards toward -1/2 to approximate a Nash equilibrium, aiming to reduce data sparsity issues.
(8) \textbf{AOT}~\cite{melnyk2024aot}, which applies Distributional Preference Alignment via Optimal Transport.
(9) \textbf{TR-DPO}~\cite{gorbatovski2024tr_dpo}, which adds synchronization between the model and a reference model every few steps to mitigate overfitting during DPO training.
(10) \textbf{ORPO}~\cite{hong2024orpo}, a reference model-free preference optimization algorithm that uses a log odds ratio penalty appended to the NLL loss, allowing for preference-aligned fine-tuning without an additional preference alignment phase.
For all algorithms, we set the learning rate to $5e\text{-}6$ and use the hyper-parameters suggested in their corresponding paper.
Additionally, we extend these algorithms with SFT loss to analyze its impact.
The SFT model trained with the chosen responses of the reasoning preference data is also included as a baseline.

Notably, most current benchmarks lack corresponding in-distribution training samples, and the data distribution of our {\dsname} may differ from that of these benchmarks.
This discrepancy can introduce additional variability when analyzing the impact of different optimization algorithms on training results.
Therefore, we use the training and validation sets of M3CoT~\cite{chen2024m3cot} for ablation studies.

The visualization results are illustrated in Figure~\ref{fig:ablation-algo}, and the numerical results are presented in Table~\ref{tab:ablation-algorithms} and \ref{tab:appendix-ablation-main}.
We can observe that almost all preference optimization methods outperform their SFT counterpart in both the Direct and CoT settings.
However, DPO and its variants struggle to enhance the CoT reasoning abilities of the model as the resulting models exhibit trivial or no improvement when answering with CoT reasoning responses compared to direct-answer responses.
On the other hand, when combining SFT Loss with these DPO variants, all algorithms are able to improve the model's CoT reasoning abilities, demonstrating that \textit{the SFT loss is a key component for enhancing CoT reasoning abilities}.
Additionally, models trained with TR-DPO, a DPO variant that updates the reference model every few steps, perform much worse when using CoT reasoning compared to direct-answer responses.
Similarly, the model trained with ODPO, a reference-model-free method, achieves worse overall performance compared to other methods extended with SFT Loss.

These results indicate that \textit{the reference model constraint on policy updates is crucial for enhancing overall reasoning abilities, and the reference model should remain frozen during training}.
Notably, models trained with DPO+ and BCO+ exhibit the best CoT performance among existing algorithms.
Therefore, we use DPO and BCO as the preference loss and quality loss. The resulting algorithm (\ie, MPO) further improves the overall performance.

\subsection{Effects on Text-only Performance}

We evaluate the text-only performance of our models on a series of benchmarks~\cite{hendrycks2020mmlu,zhang2023gaokaobench,joshi2017triviaqa,kwiatkowski2019nq,sun2020c3,lai2017race,suzgun2022bbh,cobbe2021gsm8k,hendrycks2021math,chen2023theoremqa,zhou2023ifeval,chen2021humaneval,austin2021mbpp} and report the average performance across them.
As shown in Table~\ref{tab:ablation-text-only-performance}, although our {\dsname} dataset does not include any text-only data, the MPO-trained model achieves superior average performance on these benchmarks compared to the baseline model.
The most significant improvements are observed on TheoremQA and IFEval.
Specifically, our model trained with MPO achieves an accuracy of {20.8} on TheoremQA, a benchmark consisting of complex science problems, outperforming the baseline model by 5.2 points and the SFT counterpart by 5.0 points.
Additionally, since our dataset considers responses that fail to follow instructions as negative samples when constructing data using our correctness-based pipeline, our model also exhibits enhanced instruction-following abilities on IFEval, outperforming the baseline model by 4.1 points and the SFT counterpart by 2.8 points.

\section{Conclusion}
\label{sec:conclusion}

In this work, we introduce a preference optimization (PO) process to enhance the multimodal reasoning capabilities of MLLMs.
On the data side, we design an automated pipeline for preference data construction, which is applicable to instructions both with and without clear ground truths.
Using this pipeline, we create {\dsname}, a high-quality, large-scale multimodal reasoning preference dataset.
On the model side, we propose a simple yet effective method called Mixed Preference Optimization (MPO). This algorithm aims to learn the relative preference between pairs of responses, the absolute quality of individual responses, and the process for generating preferred responses. 
The resulting models, {\modelname} and InternVL2-76B-MPO, exhibit enhanced multimodal reasoning ability and fewer hallucinations compared to their counterparts before MPO. Notably, {\modelname} even achieves performance comparable to the 10$\times$ larger InternVL2-76B on MathVista.
We hope this study could inspire further advancements in MLLMs.

{
    \small
    \bibliographystyle{ieeenat_fullname}
    \bibliography{main}
}

\clearpage
\setcounter{page}{1}
\maketitlesupplementary

\section{Implementation Details}
\label{sec:methods-details}

During the construction of samples with clear ground truths, we sample at most $32$ reasoning processes and construct at most $15$ preference pairs for each query.
When constructing data using DropoutNTP, we truncate the original response by half and ask InternVL2-8B to complete the response without the image input. Our ablation studies in Section~\ref{sec:ablation-dropntk} show that truncating the original response by 25\% or 75\% has negative effects on the final performance.
We set the temperature to $1.0$ during sampling to ensure response diversity. Besides, the maximum tiles for dynamic resolution are set to $6$ for the general VQA domain and $12$ for OCR-, document-, and chart-related domains.

During the MPO process, the global batch size is set to $256$ during training. We employ the AdamW optimizer~\cite{loshchilov2017adamw} with the $\beta_1$ of $0.9$, the $\beta_2$ of $0.999$, and the weight decay of $0.05$. The learning rate is initialized as $5e\text{-}6$. The training phases include a linear warmup that lasts until the first 5\% of training steps.
The warmup is followed by a cosine decay strategy with a minimum learning rate of 0.
The KL penalty coefficient $\beta$ is set to $0.1$.
For the Equation~\ref{eqn:final_loss}, we set $w_p$ to $0.8$, $w_q$ to $0.2$, and $w_g$ to $1$.
The model is initialized from InternVL2-8B~\cite{chen2024internvl_1_5}, and all parameters are trainable during training.
We train the model for 1 epoch.

\section{More Ablation Studies}

\subsection{Ablation Studies about PO Variants}

In this section, we present the numerical experimental results of ablation studies on the effects of different preference optimization algorithms in Table~\ref{tab:ablation-algorithms}.
We define $\Delta$ as the performance gap between CoT reasoning responses and direct-answer responses to quantitatively assess the effects of different preference optimization algorithms on CoT reasoning abilities.
Our results indicate that introducing an additional SFT loss can significantly improve the CoT performance compared to each algorithm's vanilla counterpart.

Note that, to reduce computational costs, we only extend the DPO variants, which exhibit superior performance in Table~\ref{tab:ablation-algorithms} compared to DPO, with SFT Loss.
Furthermore, the comparison of different loss combinations is presented in Table~\ref{tab:suppl-loss-combination}.
We can observe that our MPO outperforms other methods.
Notably, the model trained with DPO+cDPO performs worse than the one trained with DPO alone.
After adding generation loss, the performance of DPO+cDPO improves greatly, but it still lags behind MPO.

In addition to the ablation studies based on M3CoT, we also present the performance of models trained with DPO+ and BCO+ using our {\dsname}, as shown in Table~\ref{tab:appendix-ablation-main}.
The experimental results show that the model trained with MPO exhibits superior overall performance compared to those trained with DPO+ and BCO+.

\begin{table}[t]
\centering

\begin{tabular}{@{}lccc@{}}
\toprule
\textbf{Method} & \textbf{Direct} & \textbf{CoT} & \textbf{$\Delta$} \\
\midrule
InternVL2-8B    & 59.3            & 57.0         & \cellcolor{red!15}{-2.3}           \\
SFT             & 65.7            & 68.5         & \cellcolor{green!15}{+2.8}            \\
\midrule
DPO~\cite{rafailov2024dpo}             & 75.8            & 72.7         & \cellcolor{red!15}{-3.1}           \\
RSO~\cite{liu2023rso}             & 74.2            & 74.3         & \cellcolor{green!15}{+0.1}            \\
IPO~\cite{azar2024ipo}             & 72.8            & 73.1         & \cellcolor{green!15}{+0.3}            \\
cDPO~\cite{mitchell2023cdpo}            & 76.2            & 76.8         & \cellcolor{green!15}{+0.6}            \\
RobustDPO~\cite{chowdhury2024robustdpo}       & 75.1            & 74.2         & \cellcolor{red!15}{-0.9}           \\
BCO~\cite{jung2024bco}             & 78.1            & 78.4         & \cellcolor{green!15}{+0.3}            \\
SPPO~\cite{wu2024sppo}            & 66.2            & 67.4         & \cellcolor{green!15}{+1.2}            \\
AOT~\cite{melnyk2024aot}             & 76.7            & 76.0         & \cellcolor{red!15}{-0.7}           \\
TR-DPO~\cite{gorbatovski2024tr_dpo}          & 75.9            & 66.8         & \cellcolor{red!15}{-9.1}           \\
\midrule
ORPO~\cite{hong2024orpo}            & 66.6            & 73.9         & \cellcolor{green!15}{+7.3}            \\
DPO+       & 76.4            & 78.9         & \cellcolor{green!15}{+2.5}            \\
cDPO+      & 71.6            & 74.2         & \cellcolor{green!15}{+2.7}            \\
RobustDPO+ & 76.5            & 78.0         & \cellcolor{green!15}{+1.5}            \\
BCO+       & 77.4            & 78.4         & \cellcolor{green!15}{+1.0}            \\
AOT+       & 76.3            & 78.0         & \cellcolor{green!15}{+1.7}            \\
\midrule
MPO (ours)
& \textbf{77.7}            & \textbf{79.1}         & \cellcolor{green!15}{+1.4}            \\
\bottomrule
\end{tabular}

\caption{
\textbf{Results of models trained with different preference optimization algorithms on M3CoT.}
$\Delta$ represents the performance gap between CoT responses and direct-answer responses.
The algorithm X extended with the SFT loss is referred to as X+ for brevity.
For example, DPO+ is the combination of DPO and the SFT loss.
Note that ORPO is also equipped with the SFT loss.
}
\label{tab:ablation-algorithms}

\end{table}

\begin{table}[t]
\renewcommand{\arraystretch}{0.85}
\scriptsize
\center

\begin{tabular}{@{}lcccccc@{}}
\toprule
\textbf{Setting} & \textbf{DPO} & \textbf{\begin{tabular}[c]{@{}c@{}}BCO w.\\ DPO\end{tabular}} & \textbf{\begin{tabular}[c]{@{}c@{}}BCO w.\\ cDPO\end{tabular}} & \textbf{\begin{tabular}[c]{@{}c@{}}DPO w.\\ cDPO\end{tabular}} & \textbf{\begin{tabular}[c]{@{}c@{}}DPO w.\\ cDPO+\end{tabular}} & \textbf{\begin{tabular}[c]{@{}c@{}}MPO\\ (ours)\end{tabular}} \\ \midrule
Direct           & 75.8         & 77.5                                                          & 76.1                                                           & 76.6                                                           & 77.2                                                            & 77.7                                                          \\
CoT              & 72.7         & 78.5                                                          & 77.8                                                           & 67.1                                                           & 78.4                                                            & 79.1                                                          \\ \bottomrule
\end{tabular}
\caption{
\textbf{Comparison of different loss combinations on M3CoT.}
}
\label{tab:suppl-loss-combination}
\end{table}

\begin{table*}[t]
\centering
\setlength\tabcolsep{2.5pt}
\renewcommand{\arraystretch}{0.9}

\begin{tabular}{@{}lcccccccc@{}}
\toprule
\multirow{2}{*}{\textbf{Model Name}} & \multicolumn{3}{c}{\textbf{Reasoning}} & \multicolumn{2}{c}{\textbf{General VQA}} & \multicolumn{3}{c}{\textbf{Hallucination Evaluation}} \\ \cmidrule(l){2-4} \cmidrule(l){5-6} \cmidrule(l){7-9}
                                     & \textbf{M3CoT} & \textbf{MathVista} & \textbf{MathVision} & \textbf{MMVet} & \textbf{LLaVA-Bench} & \textbf{POPE} & \textbf{CRPE} & \textbf{MMHalBench} \\ \midrule
InternVL2-8B        & 59.3           & 58.3               & 20.4                                                               & 54.2           & 73.2                 & 86.9          & 75.0          & 3.3                 \\
InternVL2-8B-DPO+    & 80.4           & 66.4               & 23.4                                                               & 58.3           & 74.1                 & 87.6          & 75.5          & 3.4                 \\
InternVL2-8B-BCO+   & 79.6           & 66.1               & 18.8                                                               & 55.5           & 78.6                 & 88.5          & 75.5          & 3.5                 \\
\midrule
{\modelname} (ours)
& 79.2           & 67.0               & 25.7                                                               & 56.2           & 76.7                 & 88.1          & 75.4          & 3.5                 \\ \bottomrule
\end{tabular}

\caption{
\textbf{Results of models trained with DPO+, BCO+ and MPO using our {\dsname}.}
The model trained with MPO exhibits superior overall performance compared to those trained with DPO+ and BCO+.
}
\label{tab:appendix-ablation-main}

\end{table*}

\subsection{Ablation Studies on DropoutNTP}
\label{sec:ablation-dropntk}

Here, we present the ablation results for the Dropout Ratio (DR) in our proposed DropoutNTP. By default, we set DR to $0.5$, which means that we truncate the positive response by half.
Notably, setting DR to $0.25$ means using the first quarter of the positive responses for continuation.
Following the experimental settings in Section~\ref{sec:exp-ablation-rlaifv}, we replace the negative samples in RLAIF-V with the completions based on different dropout ratios.

As shown in Table~\ref{tab:ablation-sr}, the model trained with data generated using a DR of $0.75$ performs the worst. We attribute this to the fact that, with the first three-quarters of the prefix being identical, the difference in quality between the chosen and rejected responses becomes less apparent, reducing training effectiveness.
Additionally, the model trained with a DR of $0.25$ performs worse than that trained with a dropout ratio of $0.5$. We believe this is because the majority of the content in the rejected responses is generated without image input, resulting in noticeably lower quality compared to the chosen responses, which similarly hampers the training effectiveness.
Therefore, we set the DR to $0.5$.

\begin{table}[t]
\centering
\setlength\tabcolsep{5pt}
\renewcommand{\arraystretch}{0.9}

\begin{tabular}{@{}lcccc@{}}
\toprule
\multirow{2}{*}{\textbf{Method}}                                         & \multicolumn{2}{c}{\textbf{Object HalBench}} & \multicolumn{2}{c}{\textbf{MM HalBench}} \\ \cmidrule(l){2-5} 
                                                                         & \textbf{Resp. (↓)}    & \textbf{Ment. (↓)}   & \textbf{Score}    & \textbf{Hall. (↓)}   \\ \midrule
DR=0.25 & 9.3                   & 4.8                  & 3.3               & 40.6                 \\
DR=0.50 & 7.6                   & 4.1                  & 3.6               & 31.3                 \\
DR=0.75 & 11.6                  & 6.2                  & 3.3               & 36.5                 \\ \bottomrule
\end{tabular}

\caption{
\textbf{Results of DropoutNTP with different Dropout Ratios (DR).}
The model trained on samples generated with a DR value of $0.50$ achieves the best performance.
}
\label{tab:ablation-sr}

\vspace{-3mm}

\end{table}

\subsection{Effects of Data Scale.}

To evaluate the effects of the data scale, we train the model with different amounts of preference reasoning data sampled from M3CoT~\cite{chen2024m3cot}.
The M3CoT training set contains 7,861 samples annotated with corresponding rationales. To control the data volume, we adjust the maximum number of preference pairs generated for each sample, resulting in datasets of different sizes: 10K, 40K, 70K, and 100K.

As illustrated in Figure~\ref{fig:ablation-data-scale}, model accuracy consistently improves with the increasing data volume.
As the data volume rises to 100K, the model achieves its highest accuracy of 76.4 when directly answering the final answer and 78.9 when answering with CoT.
Furthermore, both the Direct and CoT performance exhibit a positive correlation between data scale and accuracy, with the CoT performance achieving higher performance across all scales.
These results highlight the importance of scaling up reasoning preference data to improve model performance.

\subsection{Effects of Hyper-parameters.}
\label{sec:exp-ablation-hyperparams}

\begin{figure}[t]
\centering
\begin{subfigure}[b]{0.21\textwidth}
    {\includegraphics[width=\textwidth]{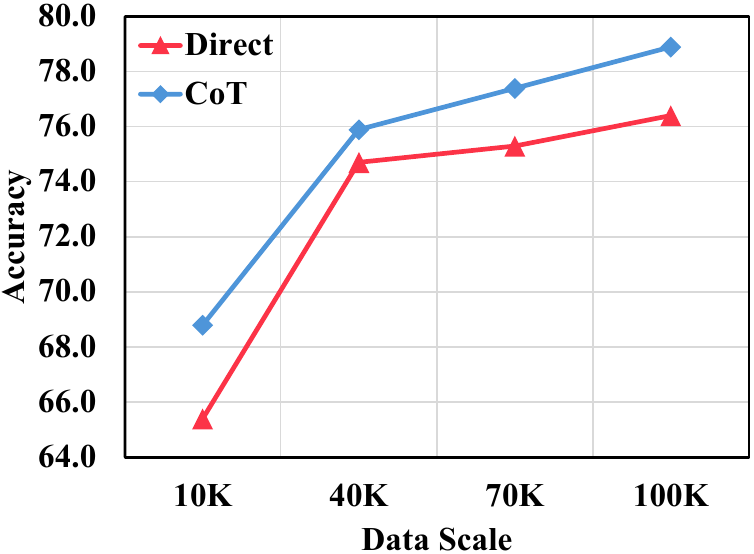}}
    \caption{}
    \label{fig:ablation-data-scale}
\end{subfigure}
\begin{subfigure}[b]{0.21\textwidth}
    {\includegraphics[width=\textwidth]{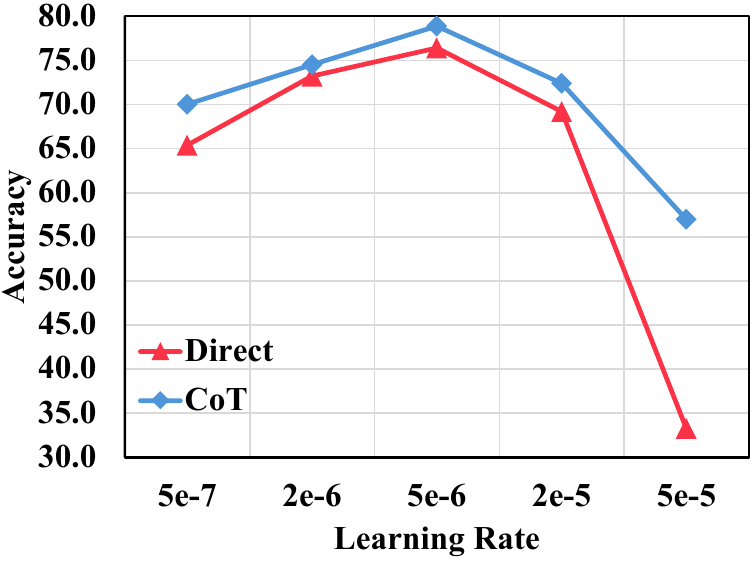}}
    \caption{}
    \label{fig:ablation-lr}
\end{subfigure}
\begin{subfigure}[b]{0.21\textwidth}
    {\includegraphics[width=\textwidth]{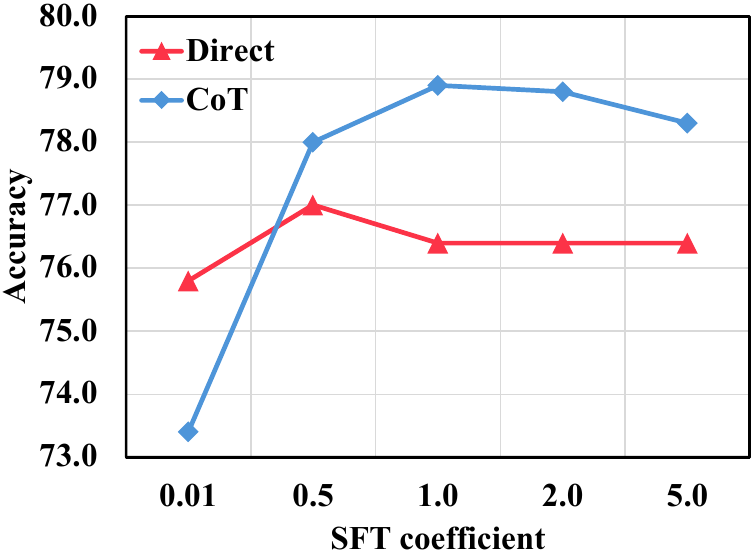}}
    \caption{}
    \label{fig:ablation-alpha}
\end{subfigure}
\begin{subfigure}[b]{0.21\textwidth}
    {\includegraphics[width=\textwidth]{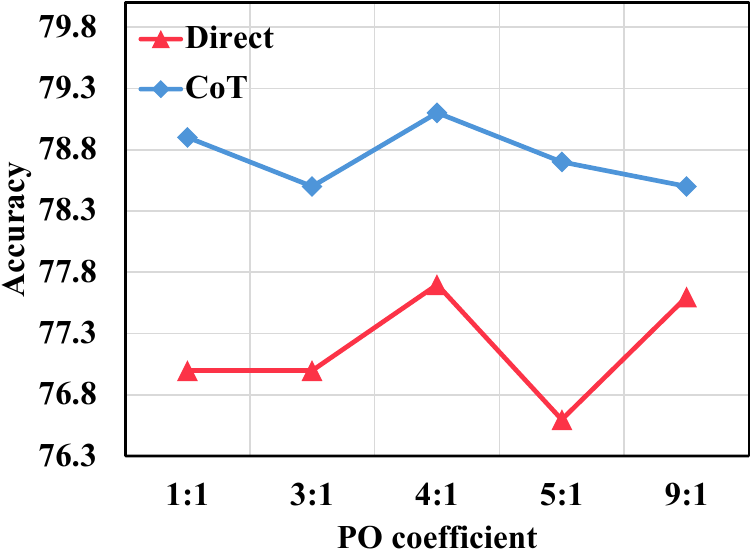}}
    \caption{}
    \label{fig:ablation-beta}
\end{subfigure}
\caption{
\textbf{Results of models trained with different data scales or hyer-parameters on M3CoT.}
The X-axis represents the corresponding data scale or hyper-parameter for this point, while the Y-axis indicates the accuracy on M3CoT.
}
\label{fig:ablation-hyperparams}

\vspace{-3mm}

\end{figure}

\begin{table*}[t]
\centering
\setlength\tabcolsep{2.5pt}
\renewcommand{\arraystretch}{0.9}

\begin{tabular}{@{}lcccccccc@{}}
\toprule
\multirow{2}{*}{\textbf{Model Name}} & \multicolumn{3}{c}{\textbf{Reasoning}} & \multicolumn{2}{c}{\textbf{General VQA}} & \multicolumn{3}{c}{\textbf{Hallucination Evaluation}} \\ \cmidrule(l){2-4} \cmidrule(l){5-6} \cmidrule(l){7-9}
                                     & \textbf{M3CoT} & \textbf{MathVista} & \textbf{MathVision} & \textbf{MMVet} & \textbf{LLaVA-Bench} & \textbf{POPE} & \textbf{CRPE} & \textbf{MMHalBench} \\ \midrule
w/o multimodal CoT
& 79.5           & 66.7               & 26.2                                                               & 54.6           & 71.4                 & 88.0          & 75.5          & 3.6                 \\
w. multimodal CoT
& 79.2           & 67.0               & 25.7                                                               & 56.2           & 76.7                 & 88.1          & 75.4          & 3.5                 \\ \bottomrule
\end{tabular}

\caption{
\textbf{Comparison of models trained on the dataset with or without multimodal CoT samples.}
The model trained on the dataset augmented with multimodal CoT samples achieve superior performance on complex Visual Question Answering (VQA) benchmarks.
}
\label{tab:appendix-ablation-cot}

\end{table*}

We conduct ablation studies on M3CoT to study the impact of the hyper-parameters, including learning rate, PO coefficient $w_p, w_q$, and SFT coefficient $w_g$.
For the PO coefficient, we control the sum of $w_p$ and $w_q$ to equal 1.0 and adjust different proportions.
Unless specifically mentioned, we set the learning rate to $5e\text{-}6$, $w_p$ to $0.8$, $w_q$ to $0.2$, and $w_g$ to $1$.
As shown in Figure~\ref{fig:ablation-lr}, the learning rate significantly affects the model's performance. 
With a relatively low learning rate of $5e\text{-}7$, the model shows moderate improvement.
As the learning rate increases to $5e\text{-}6$, the model's performance improves further, reaching optimal results across the tested learning rates and surpassing the baseline by 19.6 points.
However, further increasing the learning rate to $5e\text{-}5$ causes a drastic performance drop, suggesting that a higher learning rate may lead to overfitting or instability in training.
Additionally, the PO coefficient $w_0, w_1$ and SFT coefficient $w_2$ are crucial.
As shown in Figure~\ref{fig:ablation-alpha} and ~\ref{fig:ablation-beta}, the model achieves optimal performance with $w_p$ set to $0.8$, $w_q$ set to $0.2$, and $w_g$ set to $1$.
Notably, when $w_g$ is set to $0.01$, the performance of the CoT approach is inferior to that of directly answering the final answer, indicating the importance of the SFT Loss during the direct preference optimization.

\subsection{Effects of Multimodal Chain-of-Thought.}

In this section, we analyze the effects of multimodal CoT.
Specifically, we exclude the samples generated by the multimodal CoT methods proposed in Section~\ref{sec:methods-cot} from {\dsname}.
As shown in Table~\ref{tab:appendix-ablation-cot}, the model trained on the dataset including multimodal CoT samples (\ie, {\dsname}) achieves superior performance on complex Visual Question Answering (VQA) benchmarks.
Specifically, this model outperforms its counterpart by 1.6 points on MMVet and 5.3 points on LLaVA-Bench.
We attribute these improvements to the fact that the multimodal CoT methods not only effectively integrate multimodal information into the reasoning process but also enhance data diversity.
Furthermore, including background knowledge and visual content at the start of responses improves the quality of the negative responses generated by DropoutNTP, preventing a significant quality gap between positive and negative samples that could reduce training effectiveness.

\section{More Data Examples in {\dsname}}

\begin{figure*}[t]
\centering
\begin{subfigure}[b]{\textwidth}
    {\includegraphics[width=\textwidth]{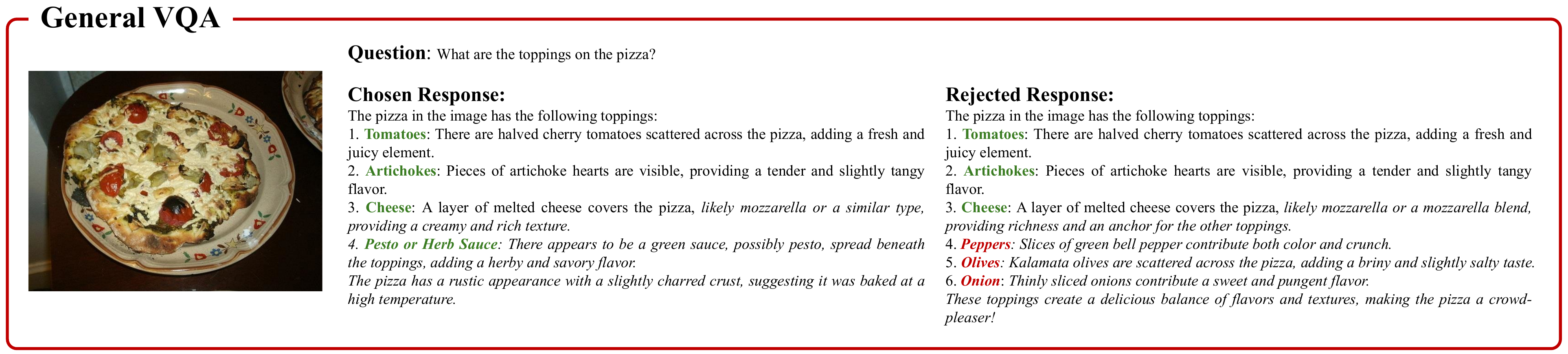}}
    \caption{}
    \label{fig:all-examples-perception-generalvqa}
\end{subfigure}
\begin{subfigure}[b]{\textwidth}
    {\includegraphics[width=\textwidth]{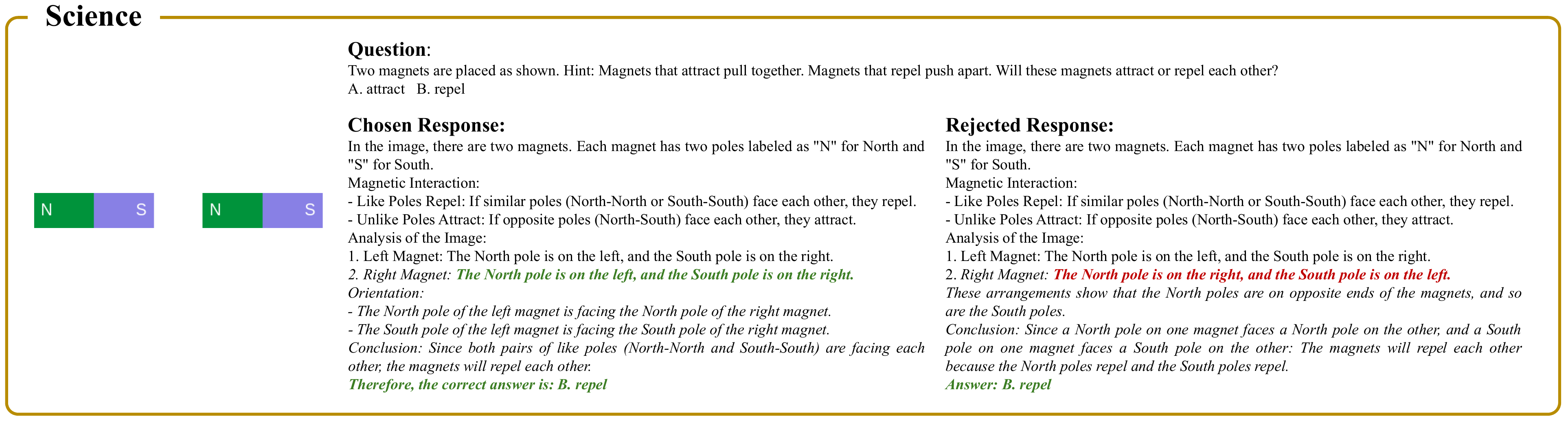}}
    \caption{}
    \label{fig:all-examples-perception-science}
\end{subfigure}
\begin{subfigure}[b]{\textwidth}
    {\includegraphics[width=\textwidth]{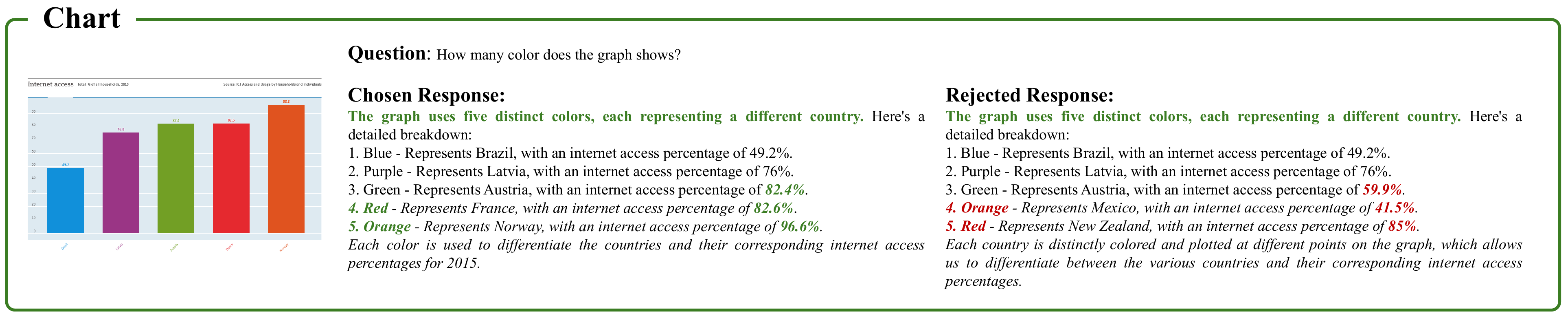}}
    \caption{}
    \label{fig:all-examples-perception-chart}
\end{subfigure}
\begin{subfigure}[b]{\textwidth}
    {\includegraphics[width=\textwidth]{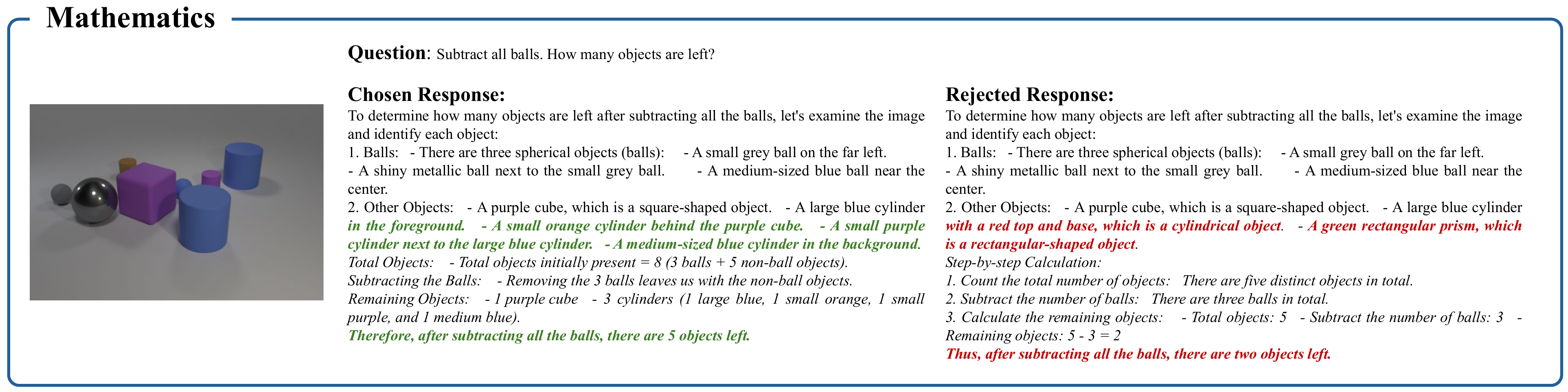}}
    \caption{}
    \label{fig:all-examples-perception-math}
\end{subfigure}
\begin{subfigure}[b]{\textwidth}
    {\includegraphics[width=\textwidth]{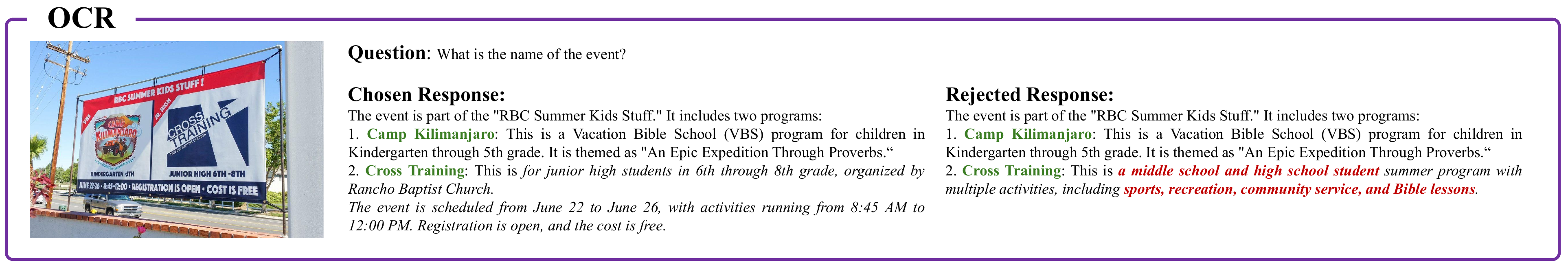}}
    \caption{}
    \label{fig:all-examples-perception-ocr}
\end{subfigure}
\end{figure*}

In this section, we provide data examples in {\dsname} for each task described in Table~\ref{tab:data-source}.
Specifically, Figure~\ref{fig:all-examples-perception-generalvqa} to \ref{fig:all-examples-perception-document} are examples from data constructed using DropoutNTP, while Figure~\ref{fig:all-examples-reasoning-science} to \ref{fig:all-examples-reasoning-math} are examples from data constructed using correctness-based pipeline.
Additionally, the examples for multimodal CoT, which is introduced in Section~\ref{sec:methods-cot}, are shown in Figure~~\ref{fig:all-examples-kcot} to \ref{fig:all-examples-gcot}.

\section{Qualitative Results}
\label{sec:suppl-qualitative}

\begin{figure*}[t]
\centering
\begin{subfigure}[b]{\textwidth}
    {\includegraphics[width=\textwidth]{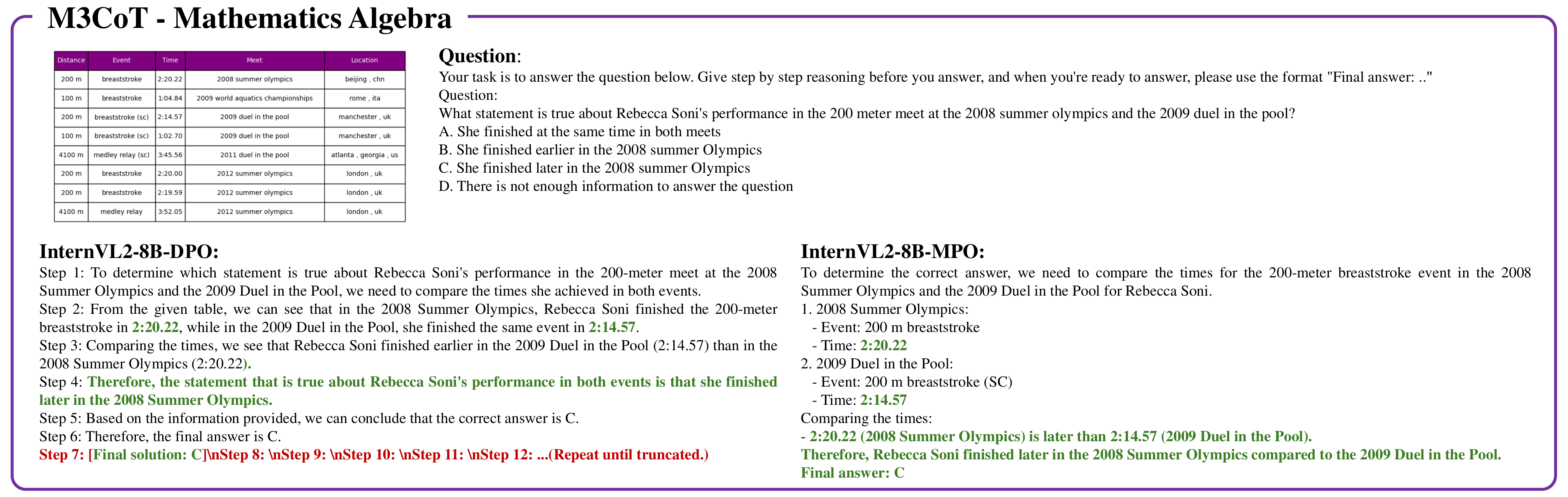}}
    \caption{}
\end{subfigure}
\begin{subfigure}[b]{\textwidth}
    {\includegraphics[width=\textwidth]{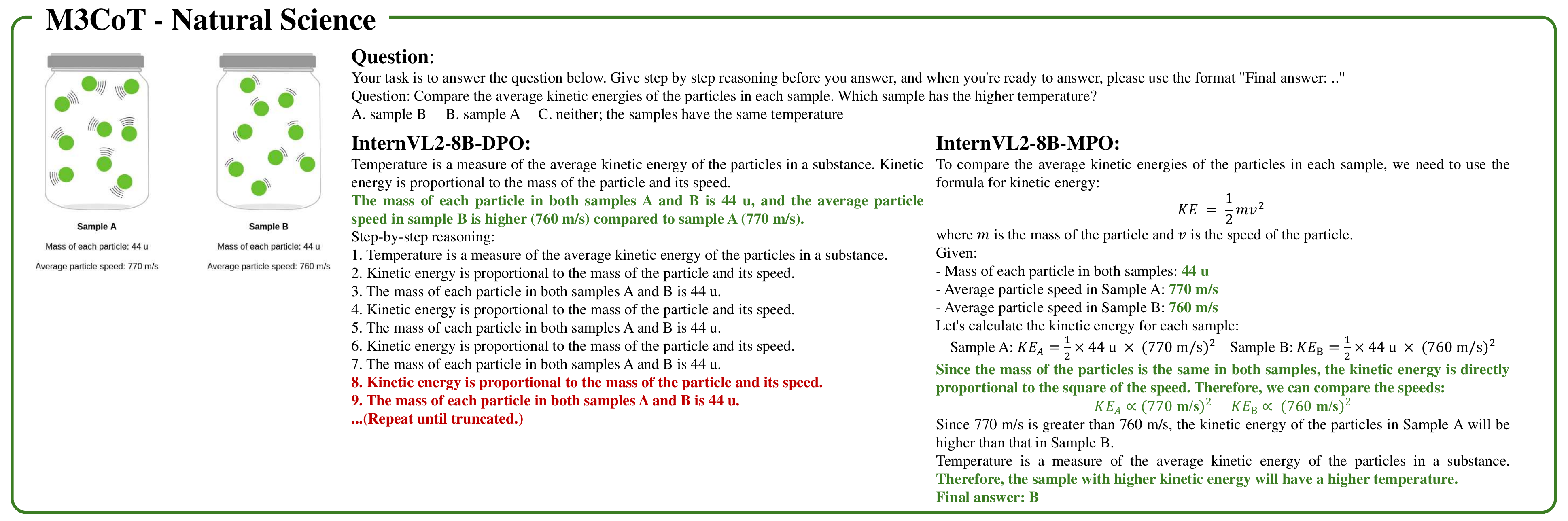}}
    \caption{}
\end{subfigure}
\begin{subfigure}[b]{\textwidth}
    {\includegraphics[width=\textwidth]{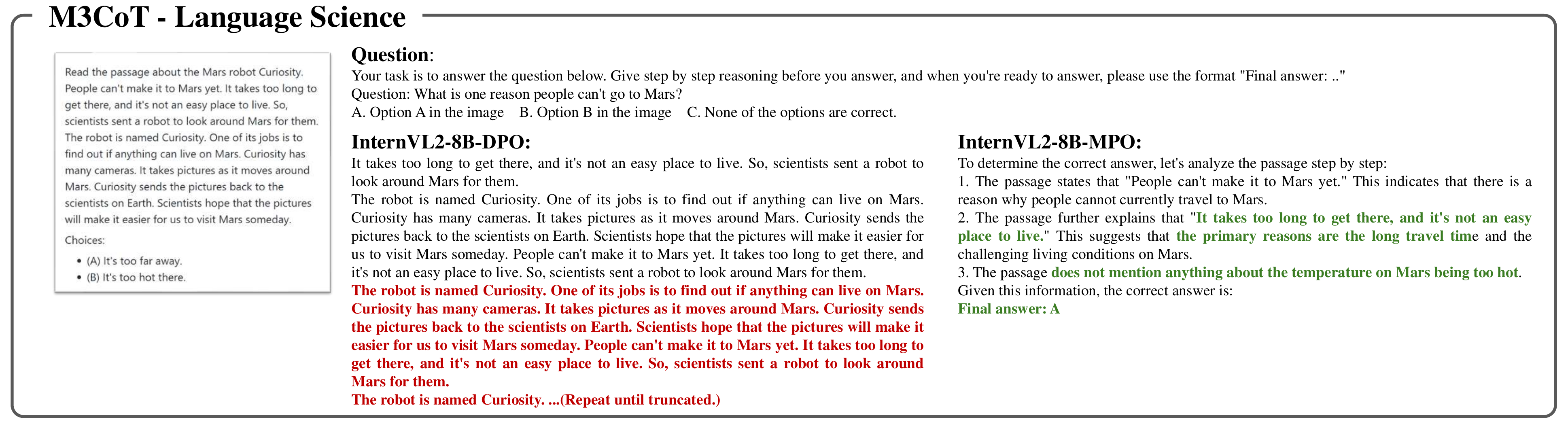}}
    \caption{}
\end{subfigure}
\caption{
\textbf{Examples of the model fine-tuned with DPO and MPO.}
\textcolor{mygreen}{Green} highlights the standout answers, while \textcolor{red}{red} highlights incorrect responses.
We observe that the model fine-tuned with DPO is more prone to generating repetitive responses.
}
\label{fig:suppl-qualitative-dpo}
\end{figure*}

\begin{figure*}[t]
\centering
\begin{subfigure}[b]{\textwidth}
    {\includegraphics[width=\textwidth]{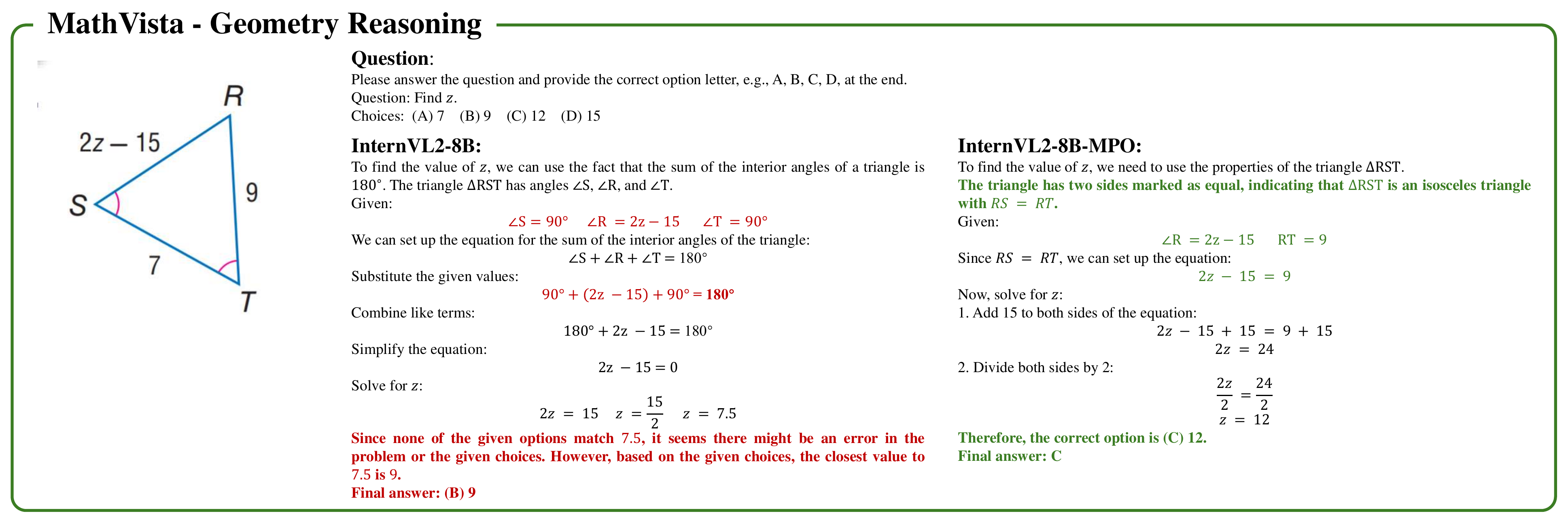}}
    \caption{}
\end{subfigure}
\begin{subfigure}[b]{\textwidth}
    {\includegraphics[width=\textwidth]{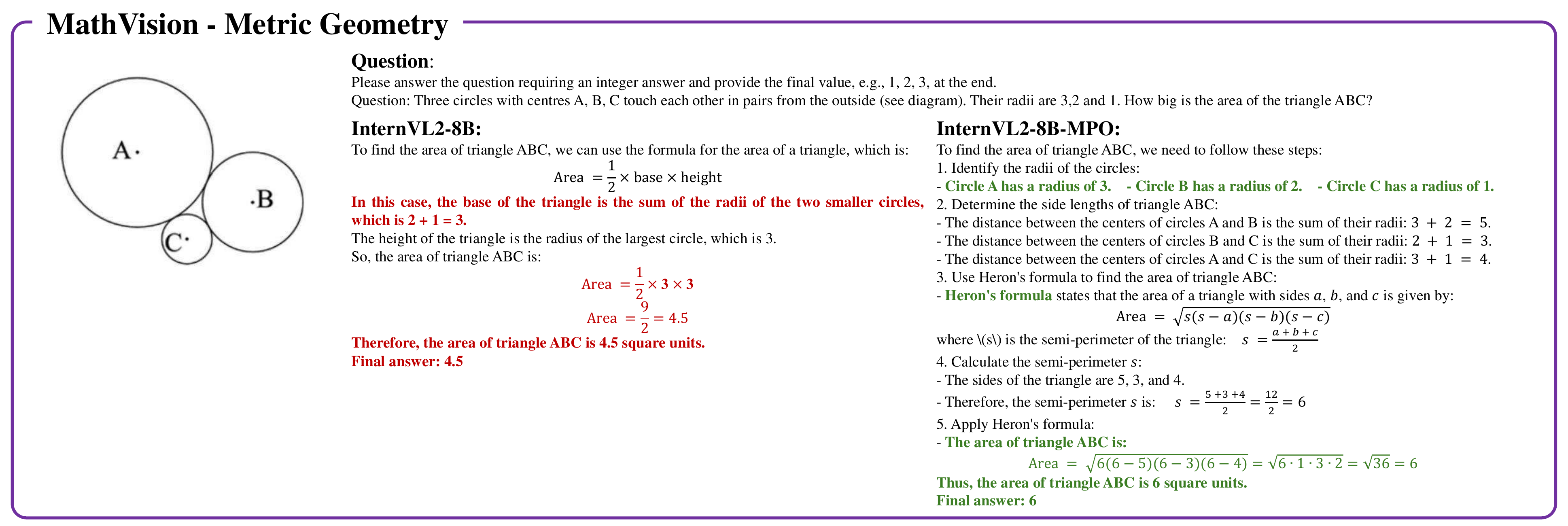}}
    \caption{}
\end{subfigure}
\caption{
\textbf{Examples of the model before and after MPO.}
\textcolor{mygreen}{Green} highlights the standout answers, while \textcolor{red}{red} highlights incorrect responses.
The model fine-tuned with MPO exhibits superior recognition and reasoning abilities.
}
\label{fig:suppl-before-and-after-mpo}
\end{figure*}

In previous sections, we evaluated our model across various
benchmarks and observed its strong performance. In this
section, we conduct a qualitative comparison between our model
and two baselines: the model before MPO (\ie, InternVL2-8B) and the model fine-tuned with DPO.
As shown in Figure~\ref{fig:suppl-qualitative-dpo}, the model fine-tuned with DPO is more prone to generating repetitive responses.
To further quantify this phenomenon, we counted the number of responses generated by InternVL2-8B-DPO that failed to produce a parsable answer. Our analysis revealed that, despite its reasonable CoT performance, 16.4\% of its responses consisted of gibberish or repetitive outputs that could not be parsed into a valid answer. In contrast, for DPO+ and MPO, only 0.4\% and 0.3\% of the responses were invalid, respectively.
Additionally, we present a qualitative comparison of the model before and after MPO in Figure~\ref{fig:suppl-before-and-after-mpo}. These examples demonstrate that our {\modelname} achieves superior performance in recognizing information from images and reasoning based on this information.

\addtocounter{figure}{-1}

\begin{figure*}[t]
\centering
\begin{subfigure}[b]{\textwidth}
    \setcounter{subfigure}{5} %
    {\includegraphics[width=\textwidth]{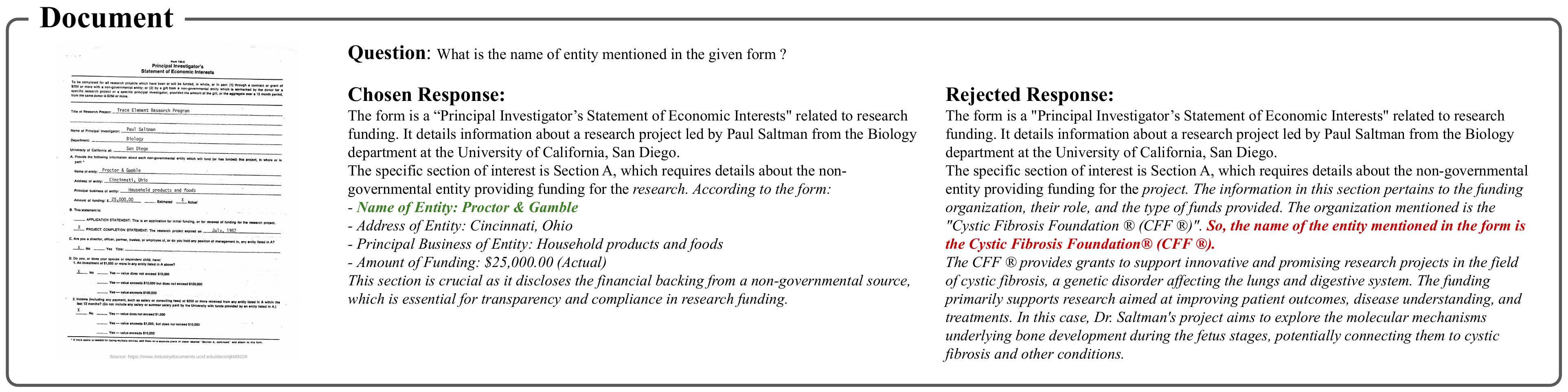}}
    \caption{}
    \label{fig:all-examples-perception-document}
\end{subfigure}
\begin{subfigure}[b]{\textwidth}
    {\includegraphics[width=\textwidth]{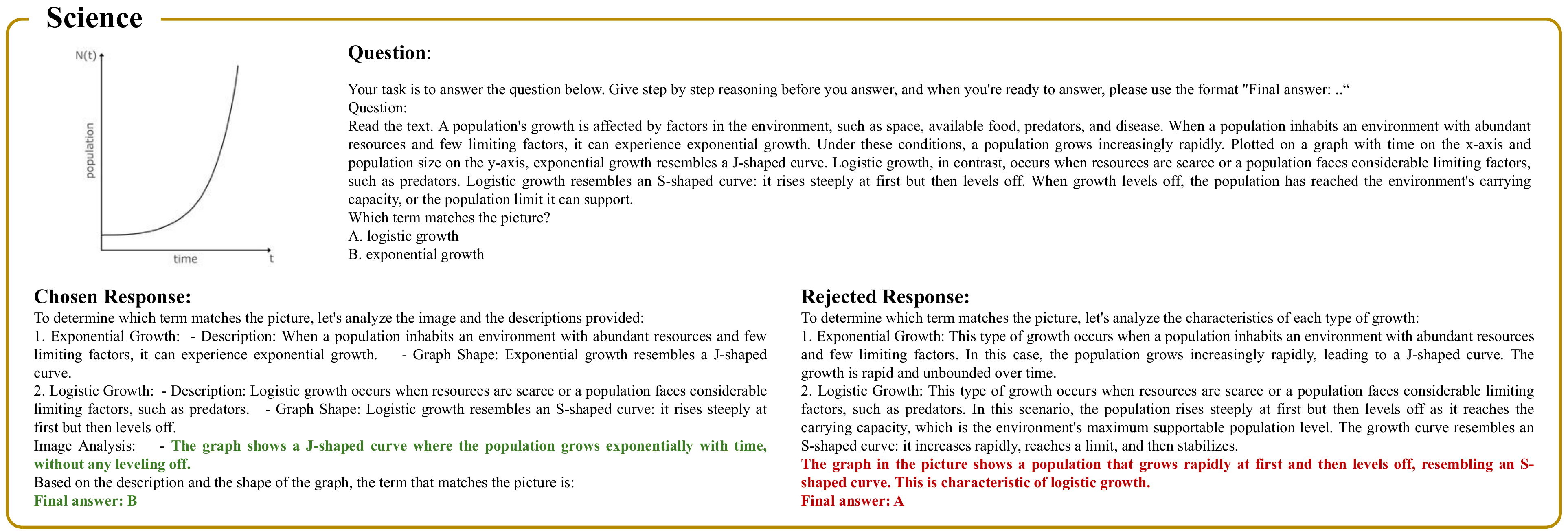}}
    \caption{}
    \label{fig:all-examples-reasoning-science}
\end{subfigure}
\begin{subfigure}[b]{\textwidth}
    {\includegraphics[width=\textwidth]{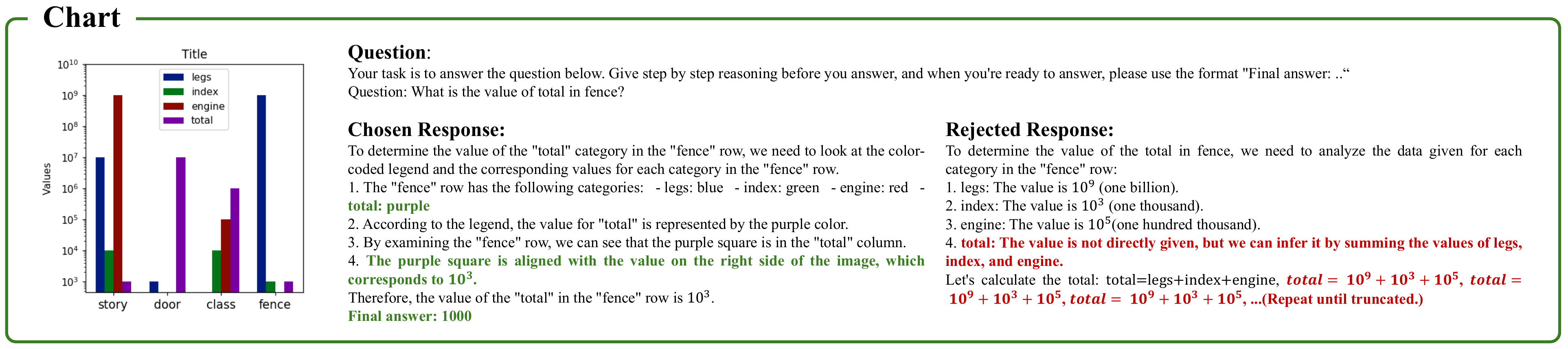}}
    \caption{}
    \label{fig:all-examples-reasoning-chart}
\end{subfigure}
\begin{subfigure}[b]{\textwidth}
    {\includegraphics[width=\textwidth]{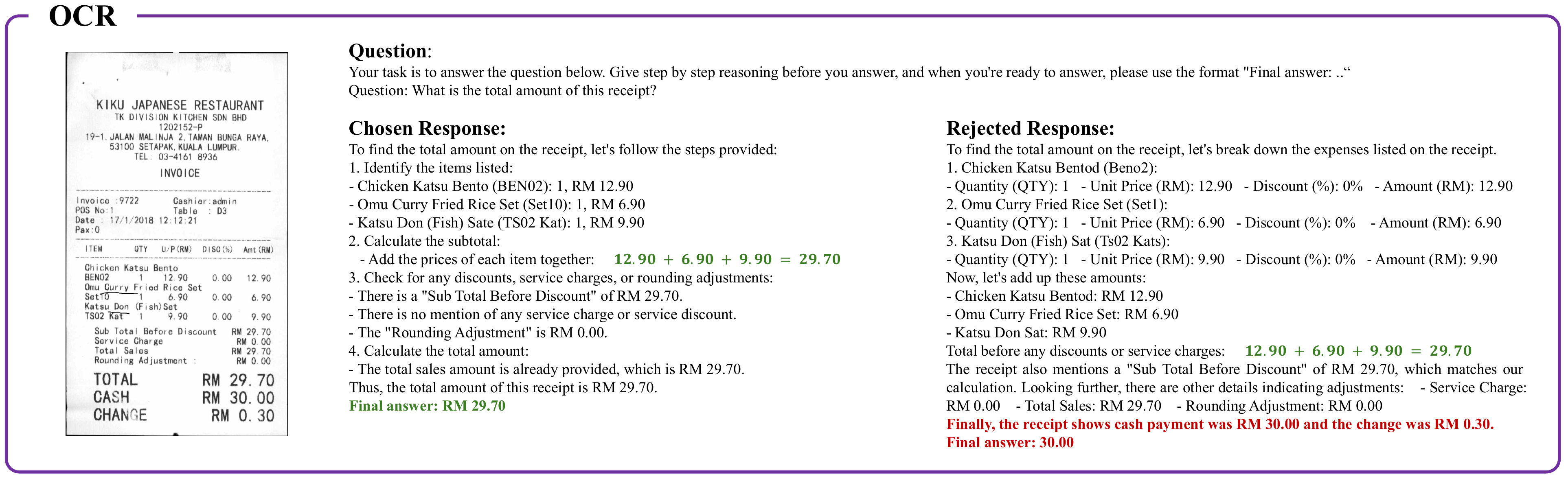}}
    \caption{}
    \label{fig:all-examples-reasoning-ocr}
\end{subfigure}
\end{figure*}

\addtocounter{figure}{-1}

\begin{figure*}[t]
\centering
\begin{subfigure}[b]{\textwidth}
    \setcounter{subfigure}{9} %
    {\includegraphics[width=\textwidth]{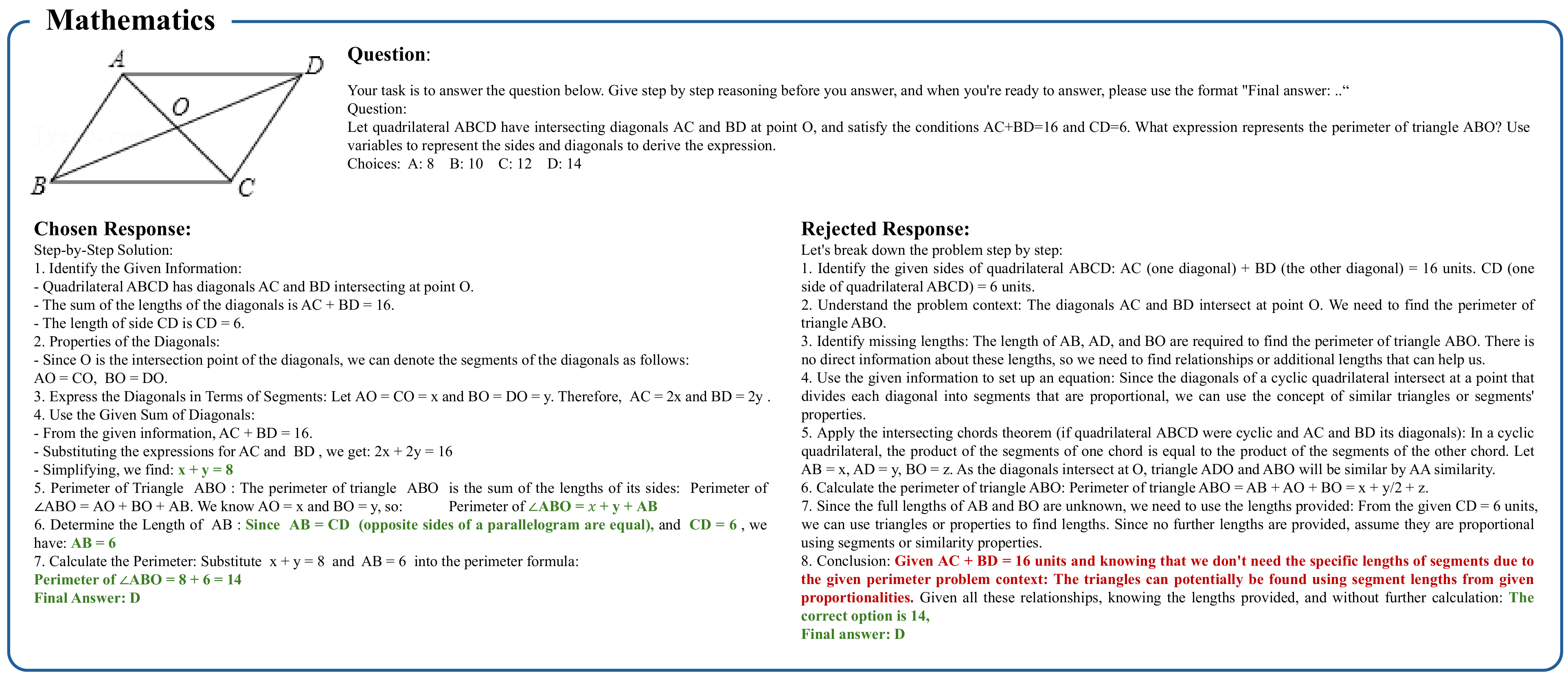}}
    \caption{}
    \label{fig:all-examples-reasoning-math}
\end{subfigure}
\begin{subfigure}[b]{\textwidth}
    {\includegraphics[width=\textwidth]{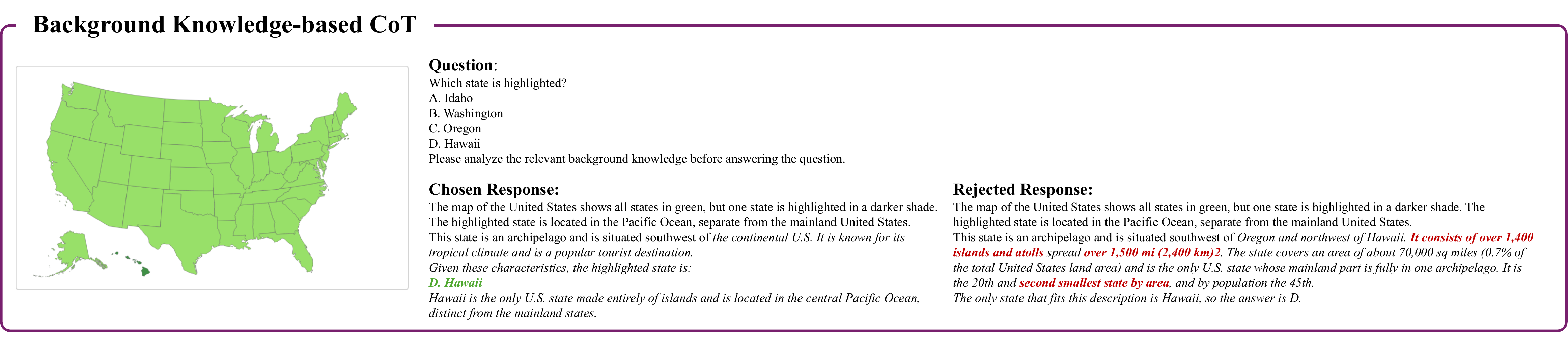}}
    \caption{}
    \label{fig:all-examples-kcot}
\end{subfigure}
\begin{subfigure}[b]{\textwidth}
    {\includegraphics[width=\textwidth]{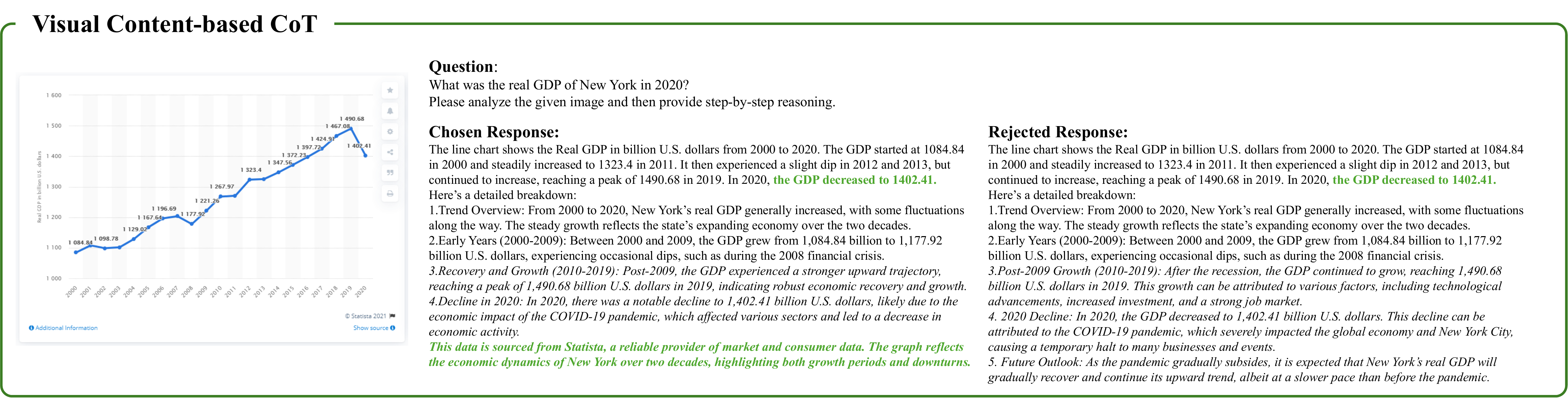}}
    \caption{}
    \label{fig:all-examples-vcot}
\end{subfigure}
\begin{subfigure}[b]{\textwidth}
    {\includegraphics[width=\textwidth]{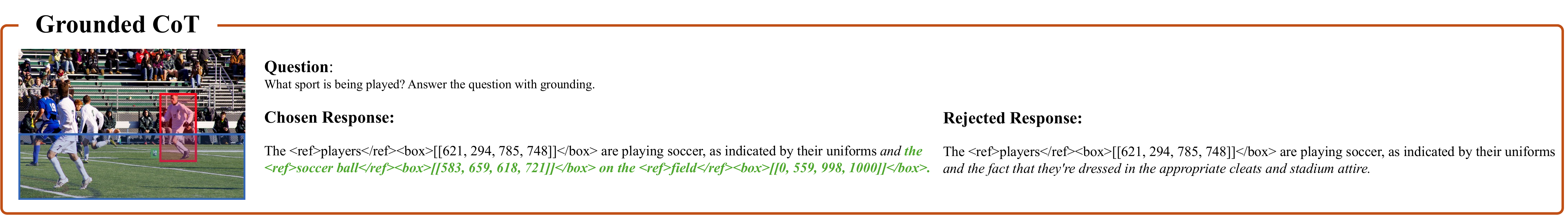}}
    \caption{}
    \label{fig:all-examples-gcot}
\end{subfigure}
\caption{
\textbf{More data examples from {\dsname}.}
Figure~\ref{fig:all-examples-perception-generalvqa} to \ref{fig:all-examples-perception-document} are examples from data constructed using DropoutNTP, while Figure~\ref{fig:all-examples-reasoning-science} to \ref{fig:all-examples-reasoning-math} are examples from data constructed using correctness-based pipeline.
Additionally, the examples for multimodal CoT, which is introduced in Section~\ref{sec:methods-cot}, are shown in Figure~~\ref{fig:all-examples-kcot} to \ref{fig:all-examples-gcot}.
}
\label{fig:more-examples-cot-reasoning-data}
\end{figure*}

\section{Discussion}
In this work, we introduce supervised fine-tuning (SFT) loss as the generation loss in Mix Preference Optimization (MPO), which may seem to contradict our original premise that SFT loss induces distribution shifts.
However, we emphasize that incorporating generation loss does not violate this premise. While generation loss and SFT loss share a similar form, their objectives are fundamentally different.

SFT training data contains only positive samples, and its goal is to teach the model to mimic high-quality response generation. As analyzed in Section~\ref{sec:intro}, this approach can lead to a distribution shift.
In contrast, Preference Optimization (PO) training data includes both positive and negative samples, with the objective of ensuring that the generation probability of positive samples is higher than that of negative samples, thereby correcting the model's response distribution. 
However, in practice, we observe that while the relative probability of generating positive samples is indeed higher than that of negative samples, the absolute probability of generating both positive and negative samples decreases simultaneously.

To address this, we introduce generation loss to ensure that while the generation probability of negative samples decreases, the probability of generating positive samples increases, thereby preventing model collapse.
Although our generation loss and SFT loss share a similar form, the generation loss primarily serves to correct the distribution, thereby mitigating the distribution shift.

\end{document}